\documentclass[10pt,twocolumn,letterpaper]{article}
\usepackage{geometry}
\usepackage{cvpr}
\usepackage{times}
\usepackage{epsfig}
\usepackage{graphicx}
\usepackage{amsmath}
\usepackage{amssymb}

\usepackage{subfigure}
\usepackage{threeparttable}
\usepackage{multirow}
\usepackage[breaklinks=true,bookmarks=false]{hyperref}
\newcommand{\revise}[1]{{\color{black}#1}}
\newcommand{\erhao}{\fontsize{21pt}{\baselineskip}\selectfont}

\cvprfinalcopy 

\def\cvprPaperID{2261} 

\ifcvprfinal\fi
\begin{document}

{\onecolumn

\noindent \textbf{\erhao{Adaptive Interaction Modeling via Graph Operations Search}}

\vspace{2cm}

\noindent {\LARGE{Haoxin Li, Wei-Shi Zheng, Yu Tao, Haifeng Hu, Jian-Huang Lai}}

\vspace{2cm}

\noindent Code is available at: \\
\ \ \ \ \ \ \ \ \ \ \ \ \url{https://github.com/lihaoxin05/graph-operations-search}

\vspace{1cm}

\noindent For reference of this work, please cite:

\vspace{1cm}
\noindent Haoxin Li, Wei-Shi Zheng, Yu Tao, Haifeng Hu and Jian-Huang Lai.
``Adaptive Interaction Modeling via Graph Operations Search.''
In \emph{Proceedings of the IEEE Conference on Computer Vision and Pattern Recognition.} 2020.

\vspace{1cm}

\noindent Bib:\\
\noindent
@inproceedings\{li2020adaptive,\\
\ \ \   title=\{Adaptive Interaction Modeling via Graph Operations Search\},\\
\ \ \  author=\{Li, Haoxin and Zheng, Wei-Shi and Tao, Yu and Hu, Haifeng and Lai, Jian-Huang\},\\
\ \ \  booktitle=\{Proceedings of the IEEE Conference on Computer Vision and Pattern Recognition\},\\
\ \ \  year=\{2020\}\\
\}
}

\restoregeometry

\title{Adaptive Interaction Modeling via Graph Operations Search}

\author{Haoxin Li\textsuperscript{1},
Wei-Shi Zheng\textsuperscript{2,3,5,}\thanks{Corresponding author},
Yu Tao\textsuperscript{2,4},
Haifeng Hu\textsuperscript{1,}\footnotemark[1],
Jian-Huang Lai\textsuperscript{2}
\\
\textsuperscript{1}{School of Electronics and Information Technology, Sun Yat-sen University, China}\\
\textsuperscript{2}{School of Data and Computer Science, Sun Yat-sen University, China}\\
\textsuperscript{3}{Peng Cheng Laboratory, Shenzhen 518005, China}\\
\textsuperscript{4}{Accuvision Technology Co. Ltd.}\\
\textsuperscript{5}{Key Laboratory of Machine Intelligence and Advanced Computing, Ministry of Education, China}\\
\tt\small lihaoxin05@gmail.com,
wszheng@ieee.org,
gytaoyu@hotmail.com \\
\tt\small huhaif@mail.sysu.edu.cn,
stsljh@mail.sysu.edu.cn
}


\maketitle
\thispagestyle{empty}

\begin{abstract}
Interaction modeling is important for video action analysis. Recently, several works design specific structures to model interactions in videos. However, their structures are manually designed and non-adaptive, which require structures design efforts and more importantly could not model interactions adaptively. In this paper, we automate the process of structures design to learn adaptive structures for interaction modeling. We propose to search the network structures with differentiable architecture search mechanism, which learns to construct adaptive structures for different videos to facilitate adaptive interaction modeling. To this end, we first design the search space with several basic graph operations that explicitly capture different relations in videos. We experimentally demonstrate that our architecture search framework learns to construct adaptive interaction modeling structures, which provides more understanding about the relations between the structures and some interaction characteristics, and also releases the requirement of structures design efforts. Additionally, we show that the designed basic graph operations in the search space are able to model different interactions in videos. The experiments on two interaction datasets show that our method achieves competitive performance with state-of-the-arts.
\end{abstract}

\section{Introduction}
Video classification is one of the basic research topics in computer vision. Existing video classification solutions can be mainly divided into two groups. The first one is the two-stream network based methods \cite{NIPS2014_5353,8454294,7780582}, which model appearance and motion features with RGB and optical flow streams respectively; the second type is the 3D convolution neural networks (CNN) based methods \cite{7410867,8099985,7940083,8237852,Tran_2018_CVPR,Luo_2019_ICCV}, which model spatiotemporal features with stacked 3D convolutions or the decomposed variants. While these methods work well on scene-based action classification, most of them obtain unsatisfactory performance on recognizing interactions, since they haven't effectively or explicitly modeled the relations.

\begin{figure}[t]
    \centering
    \includegraphics[width=1.0\linewidth]{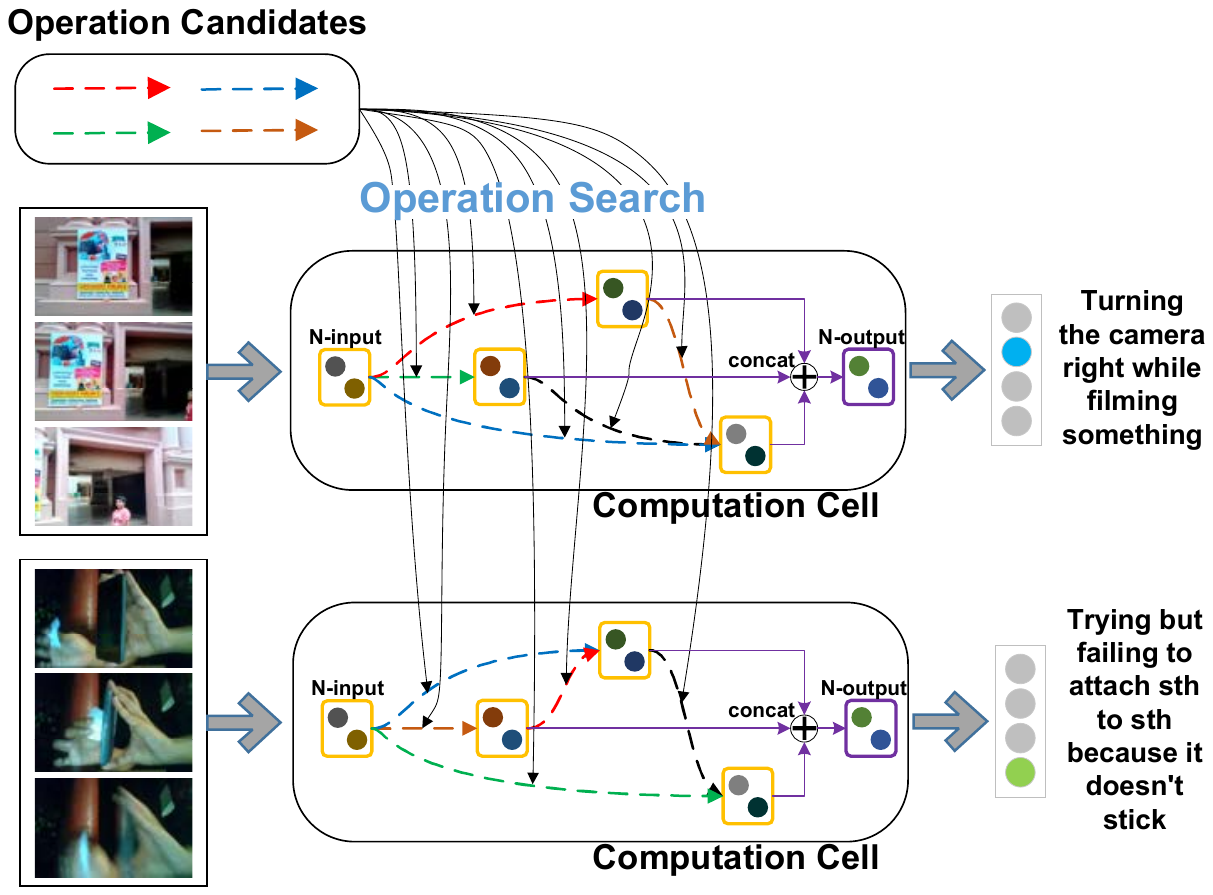}
    \caption{Illustration of our method. We search adaptive network structures to model the interactions in different videos, in which the candidate basic operations (dashed arrows) are selected (solid arrows) to construct adaptive structures for different videos.
    }
    \label{fig:concept}
\end{figure}

To model the interactions in videos, some methods employ specific structures \cite{Zhou_2018_ECCV,hussein2018timeception,jiang2019stm} to capture temporal relations. Others model the relations between entities. Non-local network \cite{8578911} and GloRe \cite{chen2018graph} design networks with self-attention and graph convolution to reason about the relations between semantic entities. CPNet \cite{Liu_2019_CVPR} aggregates features from potential correspondences for representation learning. Space-time region graphs \cite{Wang_2018_ECCV} are developed to model the interactions between detected objects with graph convolution network (GCN).

However, existing methods have to manually design network structures for interaction modeling, which requires considerable architecture engineering efforts. More importantly, the designed structures are fixed so that they could not adaptively model different interactions. For example, the two videos in Figure \ref{fig:concept} contain the interactions with greatly different complexities and properties, \ie the upper one mainly concerns the motions of the background while the lower one involves complicated relations among objects, \revise{where which kind of structures should be used to adequately model the interactions is not completely known in advance, so that it requires to construct adaptive structures for more effective interactions modeling.}

Instead of designing fixed network structures manually, we propose to automatically search adaptive network structures directly from training data, which not only reduces structures design efforts but also enables adaptive interaction modeling for different videos. As briefly illustrated in Figure \ref{fig:concept}, different operations are adaptively selected to construct the network structures for adaptive interaction modeling for different videos, which is implemented by differentiable architecture search. To construct the architecture search space, we first design several basic graph operations which explicitly capture different relations in videos, such as the temporal changes of objects and relations with the background. Our experiments show that the architecture search framework automatically constructs adaptive network structures for different videos according to some interaction characteristics, and the designed graph operations in the search space explicitly model different relations in videos. Our method obtains competitive performance with state-of-the-arts in two interaction recognition datasets.

In summary, the contribution of this paper is two-fold. (1) We propose to automatically search adaptive network structures for different videos for interaction modeling, which enables adaptive interaction modeling for different videos and reduces structures design efforts. (2) We design the search space with several basic graph operations, which explicitly model different relations in videos.

\section{Related Work}
\subsection{Action and Interaction Recognition}
In the deep learning era, action recognition obtains impressive improvements with 2D \cite{NIPS2014_5353,8454294,7780582} or 3D \cite{6165309,7410867,8237852,8099985,7940083,Tran_2018_CVPR,Luo_2019_ICCV} CNNs. 2D CNNs use RGB frames and optical flows as separate streams to learn appearance and motion representations respectively, while 3D CNNs learn spatiotemporal features with 3D convolutions or the decomposed counterparts. Some other works \cite{lin2018temporal,jiang2019stm} learn spatiotemporal representations by shifting feature channels or encoding motion features together with spatiotemporal features, which achieve high performance and efficiency. As for temporal-based actions, TRN \cite{Zhou_2018_ECCV} and Timeception \cite{hussein2018timeception} design specific structures to model the temporal relations.

To model interactions, Gupta \etal \cite{4815270} apply spatial and functional constraints with several integrated tasks to recognize interactions. InteractNet \cite{Gkioxari_2018_CVPR} and Dual Attention Network \cite{xiao2019reasoning} are proposed to model the interactions between human and objects. Some other works model the relations between entities for interaction recognition. Non-local network \cite{8578911} models the relations between features with self-attention. CPNet \cite{Liu_2019_CVPR} aggregates correspondences for representation learning. GCNs are employed to model the interactions between nodes \cite{Wang_2018_ECCV,chen2018graph}. These specific structures in the above methods are non-adaptive. In practice, however, we do not know what kinds of interactions are contained in videos, and the non-adaptive structures could not sufficiently model various interactions, which requires adaptive structures for effective modeling.

In this work, we propose to automatically search adaptive network structures with differentiable architecture search mechanism \revise{for interaction recognition.}

\subsection{Graph-based Reasoning}
Graph-based methods are widely used for relation reasoning in many computer vision tasks. For example, in image segmentation, CRFs and random walk networks are used to model the relations between pixels \cite{chandra2017dense,bertasius2017convolutional,NIPS2018_7456}. GCNs \cite{hammond2011wavelets,kipf2016semi} are proposed to collectively aggregate information from graph structures and applied in many tasks including neural machine translation, relation extraction and image classification \cite{bastings2017graph,beck2018graph,miwa2016end,wang2018zero}. Recently, GCNs are used to model the relations between objects or regions for interaction recognition. For example, Chen \etal \cite{chen2018graph} adopt GCN to build a reasoning module to model the relations between semantic nodes, and Wang \etal \cite{Wang_2018_ECCV} employ GCN to capture the relations between detected objects.

In this paper, we design the search space with basic operations based on graph. We propose several new graph operations that explicitly model different relations in videos.

\subsection{Network Architecture Search}
Network architecture search aims to discover optimal architectures automatically. The automatically searched architectures obtain competitive performance in many tasks \cite{zoph2016neural,liu2017hierarchical,zoph2018learning}. Due to the computational demanding of the discrete domain optimization \cite{zoph2018learning,real2019regularized}, Liu \etal \cite{liu2018darts} propose DARTS which relaxes the search space to be continuous and optimizes the architecture by gradient descent.

Inspired by DARTS, we employ differentiable architecture search mechanism to automatically search adaptive structures directly from training data, which facilitates adaptive interaction modeling for different videos and releases the requirement of structures design efforts.

\begin{figure*}[ht]
    \centering
    \includegraphics[width=0.95\linewidth]{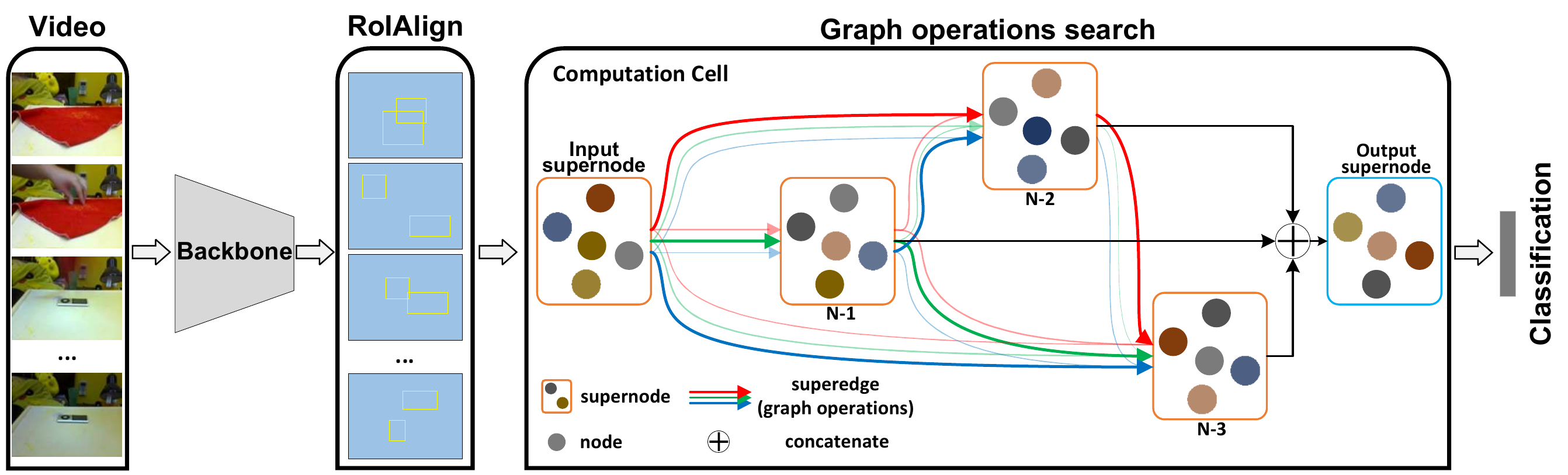}
    \caption{{\bf{Overall framework}}. Some frames are sampled from a video as the input to our model. We extract basic features of the sampled frames with a backbone CNN, and extract class-agnostic bounding box proposals with RPN model. Then we apply RoIAlign to obtain the features of proposals and regard them as node features. In the graph operations search stage, we search for a computation cell, where the supernodes are transformed by the selected graph operations on the superedges (see Section \ref{subsec:graph-operaions} and \ref{subsec:NAS} for details), to construct adaptive structures. The searched structures are used to model the interactions in the corresponding videos. Finally, the node features are pooled into a video representation for interaction recognition.}
    \label{fig:framework}
\end{figure*}

\section{Proposed Method}
In order to learn adaptive interaction modeling structure for each video, we elaborate the graph operations search method in this section. We design the architecture search space with several basic graph operations, where the candidate operations are enriched in addition to graph convolution by several proposed new graph operations modeling different relations, \eg the temporal changes and relations with background. We further develop the search framework based on differentiable architecture search to search adaptive structure for each video, which enables adaptive interaction modeling for different videos.

\subsection{Overall Framework}
We first present our overall framework for interaction recognition in Figure \ref{fig:framework}. Given a video, we sample some frames as the input to our model. We extract basic features of the sampled frames with a backbone CNN. At the same time, we extract class-agnostic RoIs for each frame with Region Proposal Network (RPN) \cite{He_2017_ICCV}. Then we apply RoIAlign \cite{He_2017_ICCV} to obtain features for each RoI. All the RoIs construct the graph for relation modeling. \revise{The nodes are exactly the RoIs, and edges are defined depending on the specific graph operations introduced in Section \ref{subsec:graph-operaions}, in which different graph operations would indicate different connections and result in different edge weights.} To obtain adaptive network structures, we employ differentiable architecture search mechanism to search adaptive structures in which graph operations are combined hierarchically. The interactions are modeled with the searched structures by transforming the node features with the selected graph operations. Finally, the output node features are pooled into a video representation for interaction classification.

In the following subsections, we describe the search space with basic graph operations and the architecture search framework in details.

\subsection{Search Space with Graph Operations}
\label{subsec:graph-operaions}
To search the network structures, we firstly need to construct a search space. We search for a computation cell to construct the network structures, as illustrated in Figure \ref{fig:framework}. A computation cell is a directed acyclic computation graph with $N$ ordered supernodes (``supernode'' is renamed from ``node'' to avoid confusion with the nodes in the graphs constructed from RoIs). \revise{Each supernode contains all the nodes and each superedge indicates the candidate graph operations transforming the node features.} In the computation cell, the input supernode is the output of the previous one, and the output is the channel-wise concatenated node features of all the intermediate supernodes.

Each intermediate supernode can be obtained by summing all the transformed predecessors (the ordering is denoted as ``N-1'', ``N-2'', ``N-3'' in Figure \ref{fig:framework}) as follows,
\begin{equation}
    \boldsymbol{X}^{(j)}=\sum_{i<j} o^{ij}(\boldsymbol{X}^{(i)}),
    \label{eqn:summing-predecessors}
\end{equation}
where $\boldsymbol{X}^{(i)}$, $\boldsymbol{X}^{(j)}$ are the node features of the $i$-th and $j$-th supernode, and $o^{ij}$ is the operation on superedge $(i,j)$. Thus the learning of cell structure reduces to learning the operations on each superedge, so that we design the candidate operations in the following.

We design the basic operations based on graph for explicit relation modeling. In addition to graph convolution, we propose several new operations, \ie \textit{difference propagation}, \textit{temporal convolution}, \textit{background incorporation} and \textit{node attention}, which explicitly model different relations in videos and serve as basic operations in the search space.

\subsubsection{Feature Aggregation}
\label{subsubsec:graph-convolution}
Graph convolution network (GCN) \cite{kipf2016semi} is commonly used to model relations. It employs feature aggregation for relation reasoning, in which each node aggregates features from its neighboring nodes as follows,
\begin{equation}
    \boldsymbol{z}_i=\delta\left(\sum_{j} {a_{ij}^f\cdot\boldsymbol{W}_f\boldsymbol{x}_j}\right),
\label{eqn:graph-convolution}
\end{equation}
where $\boldsymbol{x}_j\in\mathbb{R}^{C_{in}}$ is the feature of node-$j$ with $C_{in}$ dimensions, $\boldsymbol{W}_f\in\mathbb{R}^{C_{out}\times C_{in}}$ is the feature transform matrix applied to each node, $a_{ij}^f=\boldsymbol{x}_i^\mathsf{T}\boldsymbol{U}_f\boldsymbol{x}_j$ is the affinity between node-$i$ and node-$j$ with learnable weights $\boldsymbol{U}_f$, $\delta$ is a nonlinear activation function and the $\boldsymbol{z}_i\in\mathbb{R}^{C_{out}}$ is the updated feature of node-$i$ with $C_{out}$ dimensions. Through information aggregation on the graph, each node enhances its features by modeling the dependencies between nodes.

\subsubsection{Difference Propagation}
In videos, the differences between objects are important for recognizing interactions. But GCN may only aggregate features with weighted sum, which is hard to explicitly capture the differences. Therefore, we design an operation \textit{difference propagation} to explicitly model the differences.

\begin{figure}[t]
    \centering
    \subfigure[Difference Propagation]{
    \label{fig:diff-prop}
    \includegraphics[width=0.4\linewidth]{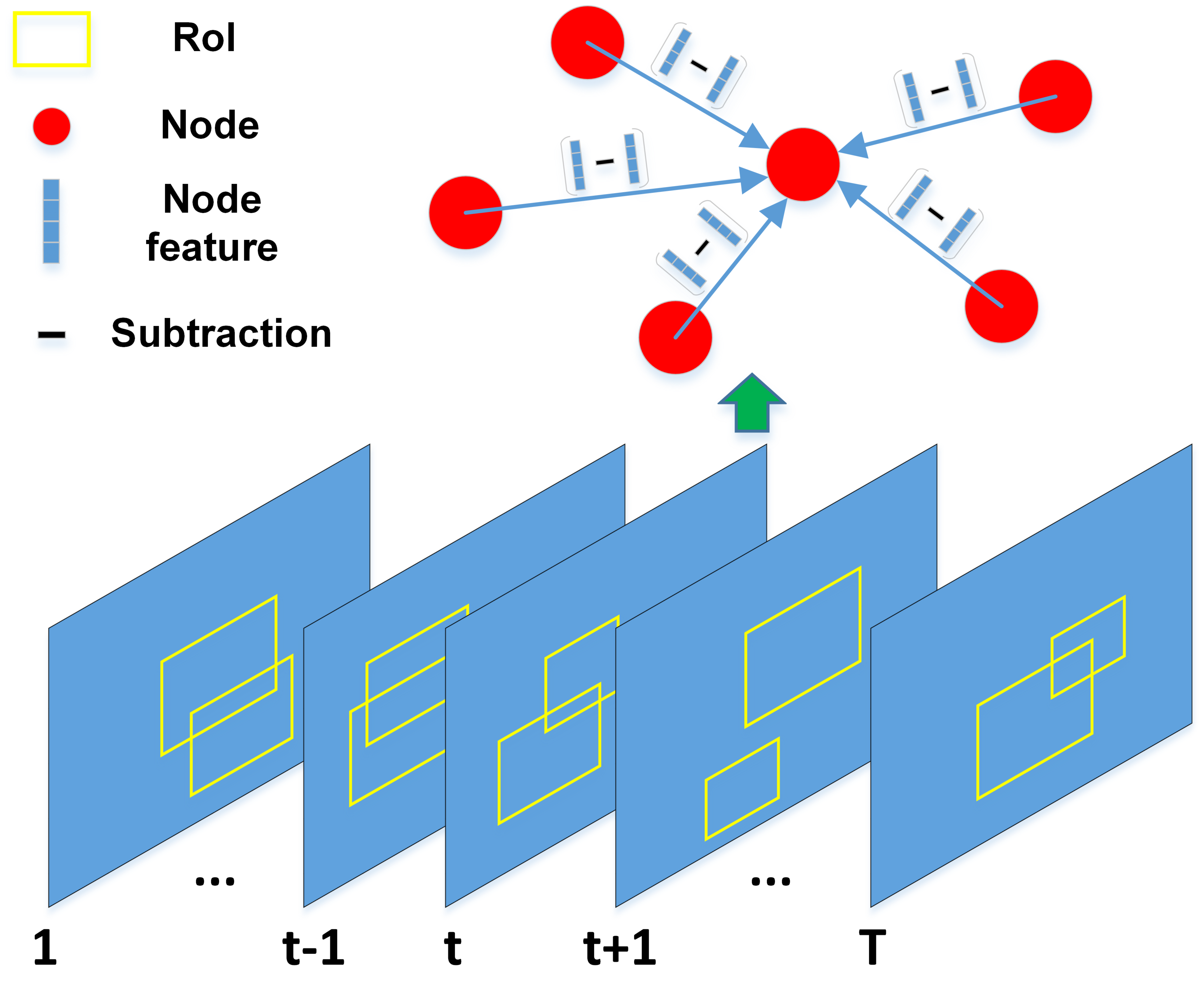}
    }
    \subfigure[Temporal Convolution]{
    \label{fig:tem-conv}
    \includegraphics[width=0.4\linewidth]{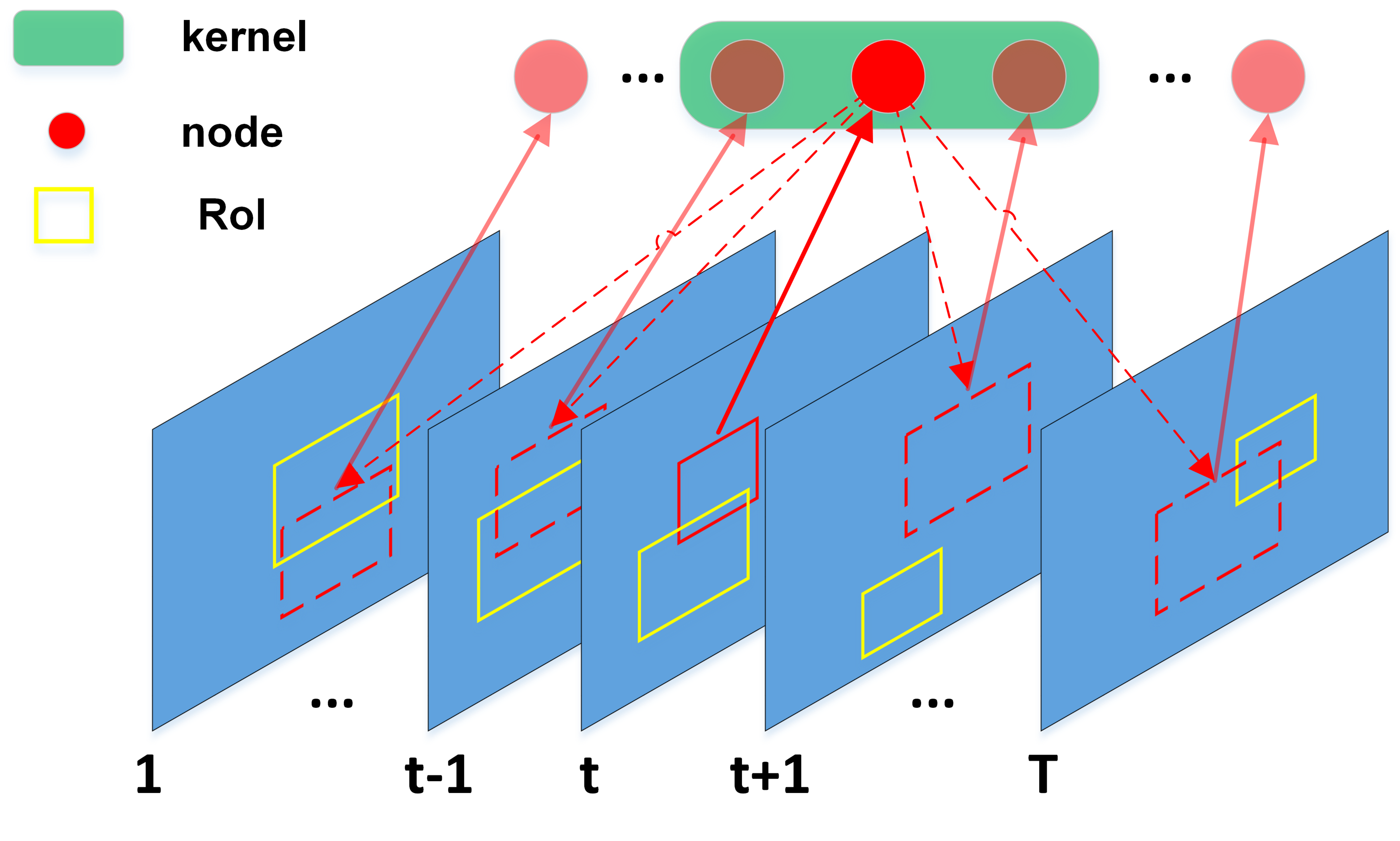}
    }
    \subfigure[Background Incorporation]{
    \label{fig:back-agg}
    \includegraphics[width=0.4\linewidth]{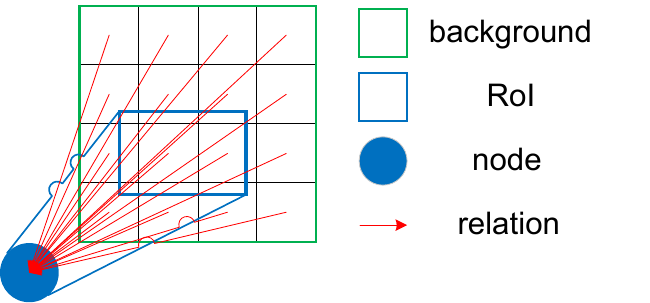}
    }
    \subfigure[Node Attention]{
    \label{fig:node-att}
    \includegraphics[width=0.4\linewidth]{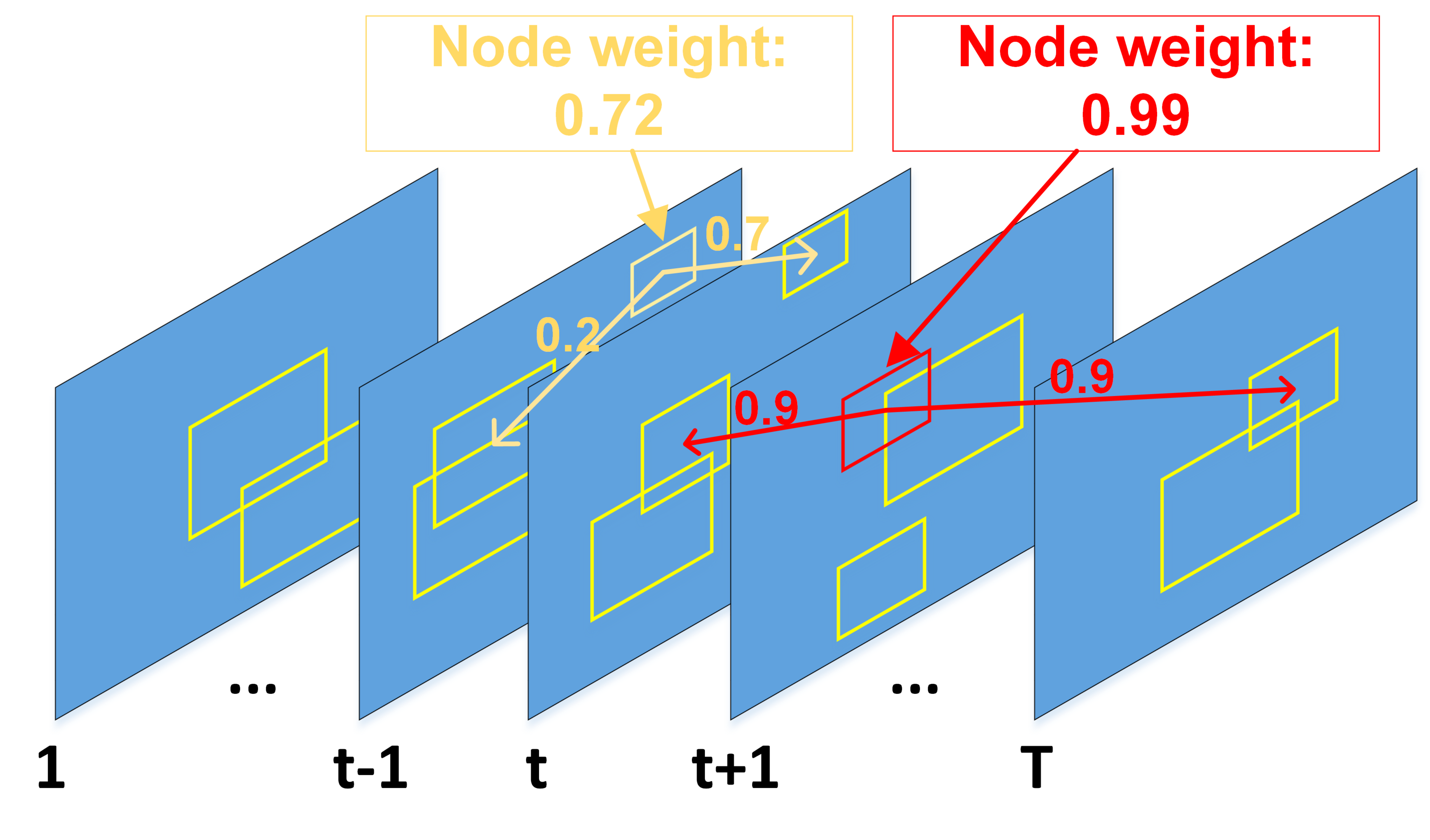}
    }
    \caption{Illustration of proposed graph operations. (a) Difference Propagation, each node propagates the differences to its neighboring nodes. (b) Temporal Convolution, each node learns temporal features with convolution over node sequences along the video. (c) Background Incorporation, each node aggregates the relations with the background. (d) Node Attention, each node learns attention weights to indicate its importance.}
\end{figure}

By slightly modifying Equation (\ref{eqn:graph-convolution}), the differences can be explicitly modeled as follows,
\begin{equation}
    \boldsymbol{z}_i=\delta\left(\sum_{j,j\neq i} {a_{ij}^d\cdot\boldsymbol{W}_d(\boldsymbol{x}_i-\boldsymbol{x}_j)}\right),
\label{eqn:difference-propagation}
\end{equation}
where the symbols share similar meanings of those in Equation (\ref{eqn:graph-convolution}). The item $(\boldsymbol{x}_i-\boldsymbol{x}_j)$ in Equation (\ref{eqn:difference-propagation}) explicitly models the differences between node-$i$ and node-$j$, and then the differences are propagated on the graph, as shown in Figure \ref{fig:diff-prop}. \textit{Difference propagation} focuses on the differences between nodes to model the changes or differences of objects, which benefits recognizing interactions relevant to the changes or differences.

\subsubsection{Temporal Convolution}
Nodes in videos are inherently in temporal orders. However, both \textit{feature aggregation} and \textit{difference propagation} model the features in unordered manners and ignore the temporal relations. Here we employ \textit{temporal convolution} to explicitly learn temporal representations.

In temporal convolutions, we firstly obtain node sequences in temporal order. Given node-$i$ in the $t$-th frame, \revise{we find its nearest node (not required to represent the same object) in each frame measured by the inner product of node features and arrange them in temporal order for a sequence,}
\begin{equation}
    \boldsymbol{X}_i=[\boldsymbol{x}^0_i,\cdots,\boldsymbol{x}^t_i,\cdots,\boldsymbol{x}^{T-1}_i],
\end{equation}
where $\boldsymbol{x}^0_i,\cdots,\boldsymbol{x}^{T-1}_i$ denote the nearest nodes in frame $0,\cdots,T-1$ with reference to the given node $\boldsymbol{x}^t_i$.

Then we conduct temporal convolutions over the node sequence as shown in Figure \ref{fig:tem-conv},
\begin{equation}
    \boldsymbol{z}_i=\delta(\boldsymbol{W}_t\ast\boldsymbol{X}_i),
\end{equation}
where $\ast$ denotes temporal convolution and $\boldsymbol{W}_t$ is the convolution kernel. The \textit{temporal convolution} explicitly learns \revise{the temporal representations to model the significant appearance changes of the node
sequence, which is essential for identifying interactions with temporal relations.}

\subsubsection{Background Incorporation}
The node features derived from RoIAlign exclude the background information. However, background is useful since the objects probably interact with the background. This inspires us to design the \textit{background incorporation} operation.

In each frame, the detected objects have different affinities with different regions in the background, as illustrated in Figure \ref{fig:back-agg}. Denote the feature of node-$i$ in the $t$-th frame as $\boldsymbol{x}_i^t\in\mathbb{R}^{C_{in}}$ and the background feature map corresponding to the $t$-th frame as $\boldsymbol{y}^t\in\mathbb{R}^{h\times w\times C_{in}}$. The affinity between $\boldsymbol{x}_i^t$ and $\boldsymbol{y}_{j}^t$ ($j=1,\cdots,h\times w$) can be calculated as $a_{ij}^b={\boldsymbol{x}_i^t}^{\mathsf{T}}\boldsymbol{U}_b\boldsymbol{y}_j^t$ with learnable $\boldsymbol{U}_b$. The $a_{ij}^b$ indicates the relations between the node and the background with spatial structure, which could be transformed into node features,
\begin{equation}
    \boldsymbol{z}_i^r=\boldsymbol{V}_b \boldsymbol{a}_i^b,
    \label{eqn:background-relation}
\end{equation}
where $\boldsymbol{a}_i^b=[a_{i1}^b;a_{i2}^b;\cdots;a_{i(h\cdot w)}^b]\in\mathbb{R}^{h\cdot w}$ is the affinity vector, and $\boldsymbol{V}_b\in\mathbb{R}^{C_{out}\times (h\cdot w)}$ is the transform matrix transforming the affinity vector into node features.

In addition, the background features can be aggregated according to the affinity $a_{ij}^b$ to model the dependencies between detected objects and the background,
\begin{equation}
    \boldsymbol{z}_i^a=\sum_{j=1,\cdots,h\times w} a_{ij}^b\cdot\boldsymbol{W}_b\boldsymbol{y}_j.
\label{eqn:background-aggregation}
\end{equation}

Finally, the updated node features are the combination of the two features above followed by a nonlinear activation,
\begin{equation}
    \boldsymbol{z}_i=\delta(\boldsymbol{z}_i^r+\boldsymbol{z}_i^a).
    \label{eqn:background-total}
\end{equation}

\subsubsection{Node Attention}
The graph contains hundreds of nodes but they contribute differently to recognizing interactions. Some nodes irrelevant to the interaction serve as outliers that interfere the interaction modeling, so it is reasonable to weaken the outliers with attention scheme.

The outliers are often the nodes wrongly detected by RPN, which usually have few similar nodes and their similar nodes do not locate regularly at specific regions or along the videos, as briefly illustrated in Figure \ref{fig:node-att}. So that we calculate the attention weights according to the similarities and relative positions to the top-$M$ similar nodes.
\begin{equation}
\begin{split}
&\boldsymbol{z}_i=w_i\cdot\boldsymbol{x}_i, \\
&w_i=\sigma(\boldsymbol{W}_n\left[\boldsymbol{a}_i^n;\Delta\boldsymbol{s}_i\right]), \\
&\boldsymbol{a}_i^n=\left[a_{ij_1}^n;a_{ij_2}^n;\cdots;a_{ij_M}^n\right], \\
&\Delta\boldsymbol{s}_i=\left[
\begin{array}{c}
     \boldsymbol{s}_i-\boldsymbol{s}_{j_1} \\
     \boldsymbol{s}_i-\boldsymbol{s}_{j_2} \\
     \cdots \\
     \boldsymbol{s}_i-\boldsymbol{s}_{j_M}
\end{array}
\right],
\end{split}
\end{equation}
where $w_i$ is the attention weight of $\boldsymbol{x}_i$, which is calculated from similarity vector $\boldsymbol{a}_i^n$ and relative positions $\Delta\boldsymbol{s}_i$, $\sigma$ is the sigmoid nonlinear function, $j_m$ is the node index of node-$i$'s $m$-th similar nodes measured by inner product, and $a_{ij_m}^n$ is the inner product of node features between node-$i$ and node-$j_m$, and $\boldsymbol{s}_i=[x_i;y_i;t_i]$ is the normalized spatial and temporal positions of node-$i$. With the attention weights, we are able to focus on informative nodes and neglect the outliers.

The graph operations above explicitly capture different relations in videos and serve as the basic operations in the architecture search space, which facilitates structure search in Section \ref{subsec:NAS}.

\subsection{Searching Adaptive Structures}
\label{subsec:NAS}
With the constructed search space, we are able to search adaptive structures for interaction modeling. We employ differentiable architecture search mechanism in DARTS \cite{liu2018darts} to develop our search framework, and revise the learning of operation weights to facilitate search of adaptive interaction modeling structures.

\noindent
{\bf{DARTS.}} DARTS utilizes continuous relaxation to learn specific operations ($o^{ij}$ in Equation (\ref{eqn:summing-predecessors})) on the superedges. The softmax combination of all the candidate operations are calculated as the representation of each supernode,
\begin{equation}
    \bar{o}^{ij}(\boldsymbol{X}^{(i)})=\sum_{o\in \mathbb{O}} \frac{exp(\alpha_o^{ij})}{\sum_{o^{'}\in\mathbb{O}} exp(\alpha_{o^{'}}^{ij})}\ o(\boldsymbol{X}^{(i)}),
    \label{eqn:DARTS-continuous}
\end{equation}
where $\mathbb{O}$ is the set of candidate operations, $o$ represents a specific operation, $\alpha_o^{ij}$ is the operation weight of operation $o$ on superedge $(i,j)$, and the $\bar{o}^{ij}(\boldsymbol{X}^{(i)})$ is the mixed output. In this way, the cell structure learning reduces to the learning of operation weights $\alpha_o^{ij}$.

To derive the discrete structure after the search procedure converges, the operation with strongest weight is selected as the final operation on superedge $(i,j)$,
\begin{equation}
    o^{ij}=\mathop{\arg\max}_{o\in\mathbb{O}}\ \alpha_{o}^{ij}.
\end{equation}

\vspace{0.1cm}
\noindent
{\bf{Adaptive Structures.}} Since the interactions differ from video to video, we attempt to learn adaptive structures for automatical interaction modeling. However, the operation weights $\alpha_{o}^{ij}$ in Equation (\ref{eqn:DARTS-continuous}) is non-adaptive. So that we modify the $\alpha_{o}^{ij}$ to be adaptive by connecting them with the input video through a fully-connected (FC) layer,
\begin{equation}\label{eqn:ada-structure}
  \alpha_o^{ij}=\boldsymbol{A}_o^{ij}\boldsymbol{X},
\end{equation}
in which $\boldsymbol{X}$ is the global feature of input video (global average pooling of the backbone feature) and $\boldsymbol{A}_o^{ij}$ is the learnable structure weights corresponding to operation $o$ on superedge $(i,j)$. In this way, adaptive structures are constructed for different videos to model the interactions.

Unlike alternatively optimizing the model in training and validation set to approximate the architecture gradients in DARTS, we jointly optimize the structure weights and the weights in all graph operations in training set to learn adaptive structures.

\noindent
{\bf{Fixing Substructures.}} It is time consuming to search stable structures with too many candidate operations. We attempt to reduce the number of basic operations by combining several operations into fixed substructures and regarding the fixed substructures as basic operations in the search space. For example, we connect \textit{feature aggregation} and \textit{node attention} sequentially into a fixed combination, and put it after the other 3 graph operations to construct 3 fixed substructures for search (as shown on the superedges in Figure \ref{fig:nas-ada-structure}).

By this means, we accelerate search by simplifying the search space and also deepen the structures because each superedge contains multiple graph operations.

\noindent
{\bf{Diversity Regularization.}} We find that the search framework easily selects only one or two operations to construct structures, because these operations are easier to optimize. However, other operations are also effective on interaction modeling, so we hope to keep more operations activated in the searched structures. We introduce the variance of operation weights as an auxiliary loss to constraint that all the operations would be selected equally,
\begin{equation}\label{eqn:var-loss}
  L_{var}=\frac{1}{\vert\mathbb{O}\vert-1}\sum_{o\in \mathbb{O}}(\alpha_o-\bar{\alpha})^2,
\end{equation}
where $\alpha_o=\sum_{(i,j)}\alpha_o^{ij}$, $\bar{\alpha}$ is the mean of $\alpha_o$. The variance loss is added to the classification loss for optimization.

\section{Experiments}
\subsection{Datasets}
We conduct experiments on two large interaction datasets, Something-Something-V1(Sth-V1) and Something-Something-V2(Sth-V2) \cite{8237884} (see Figure \ref{fig:nodes-vis} and \ref{fig:case-study} for some example frames). Sth-V1 contains 108,499 short videos across 174 categories. The recognition of them requires interaction reasoning and common sense understanding. Sth-V2 is an extended version of Sth-V1 which reduces the label noises.

\subsection{Implementation Details}
\label{subsec:impl-details}
\revise{In the training, we employ stagewise training of the backbone and the graph operations search for easier convergence. And we optimize the weights in all graph operations and the structure weights ($\boldsymbol{A}_o^{ij}$ in Equation (\ref{eqn:ada-structure})) alternately to search adaptive structures.}

\revise{In the structures search stage,} we include the \textit{zero} and \textit{identity} as additional candidate operations. Following \cite{chen2019progressive}, we add dropout after \textit{identity} to avoid its domination in the searched structures. We use 3 intermediate supernodes in each computation cell. The weight for auxiliary variance loss $L_{var}$ (Equation (\ref{eqn:var-loss})) is set to 0.1.



More details about the model, training procedure and data augmentation are included in supplementary materials.

\subsection{Analysis of Architecture Search Framework}
In this section, we analyze our architecture search framework. First we compare the interaction recognition accuracy of our searched structures with our baselines, and the results are shown in Table \ref{tab:nas-scheme-cmp}. It is observed that our searched structures obtain about 3\% improvements over the baselines, \ie \textit{global pooling} (global average pooling of the backbone feature) and \textit{pooling over RoIs} (average pooling over all the RoI features), indicating that the searched structures are effective to model interactions and improve recognition performance. In the following, we show the searched structures and analyze the effects of adaptive structures.

\begin{table}[t]
    \scriptsize
    \centering
    \begin{threeparttable}
    \begin{tabular}{c|c|c}
        \hline
        Search schemes & V1 Val\tnote{1} Acc & V2 Val\tnote{1} Acc \\
        \hline\hline
        global pooling & 48.1 & 60.3 \\
        pooling over RoIs & 48.3 & 60.3 \\
        \hline
        non-adaptive (only testing)\tnote{2} & 50.2 & 62.4 \\
        non-adaptive (training and testing)\tnote{3} & 50.8 & 63.1 \\
        adaptive & 51.4 & 63.5 \\
        \hline
    \end{tabular}
    \begin{tablenotes}
     \tiny
     \item[1] Something-Something-V1 validation set and Something-Something-V2 validation set
     \item[2] Only one searched structure (corresponding to most training videos) is used for testing.
     \item[3] The structure are non-adaptive both in training and testing.
   \end{tablenotes}
   \end{threeparttable}
    \caption{Interaction recognition accuracy (\%) comparison of different search schemes.}
    \label{tab:nas-scheme-cmp}
\end{table}

\begin{figure}[t]
    \centering
    \includegraphics[width=\linewidth]{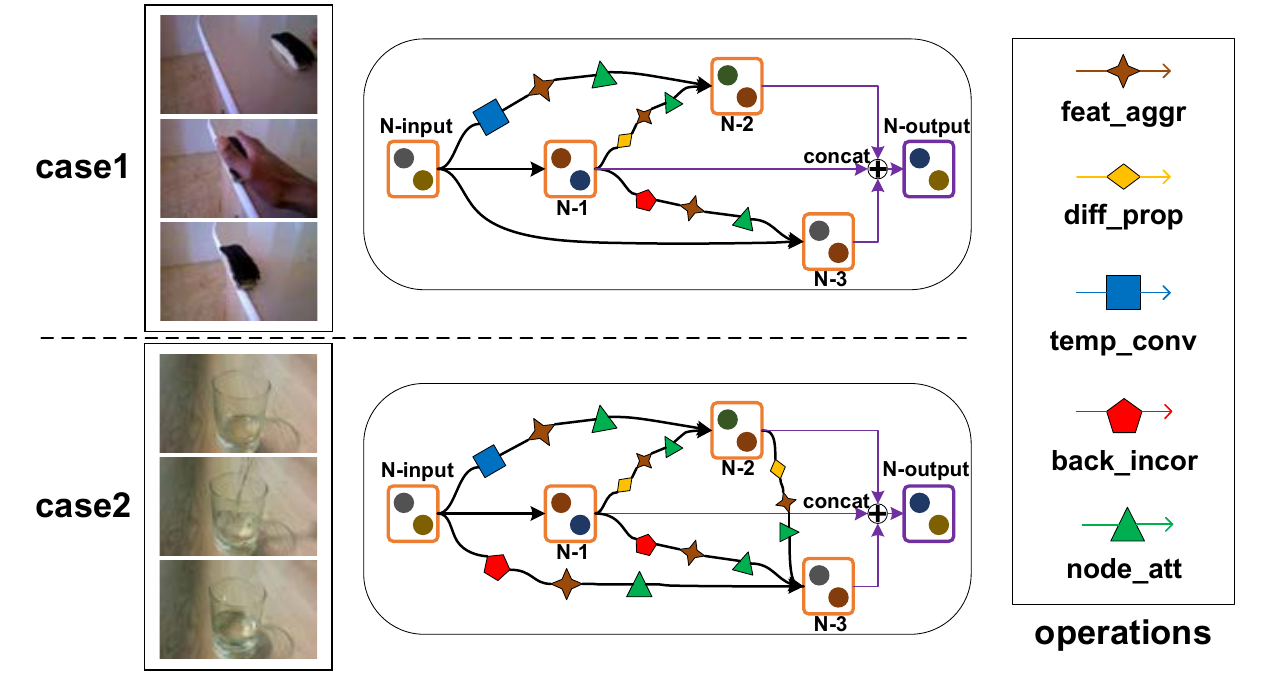}
    \caption{Two example videos and their corresponding structures. In the figure, ``feat\_aggr'', ``diff\_prop'', ``temp\_conv'', ``back\_incor'', ``node\_att'' represent \textit{feature aggregation}, \textit{difference propagation}, \textit{temporal convolution}, \textit{background incorporation} and \textit{node attention}, respectively.}
    \label{fig:nas-ada-structure}
\end{figure}

\subsubsection{Searched Structures}
Figure \ref{fig:nas-ada-structure} shows two examples of the input videos and the corresponding searched structures. From the searched structures we observe that our architecture search framework learns adaptive structures for different input videos. The main differences between the two structures are the superedges entering ``N-3'', where \textit{case1} learns simple structure but \textit{case2} selects complicated structure with more graph operations. Perhaps \textit{case2} is confusing with other interactions and requires complicated structures to capture some detailed relations for effective interaction modeling.

\begin{figure}[t]
    \centering
    \subfigure[Match and mismatch classification comparison of case 1.]{
    \includegraphics[width=0.47\linewidth]{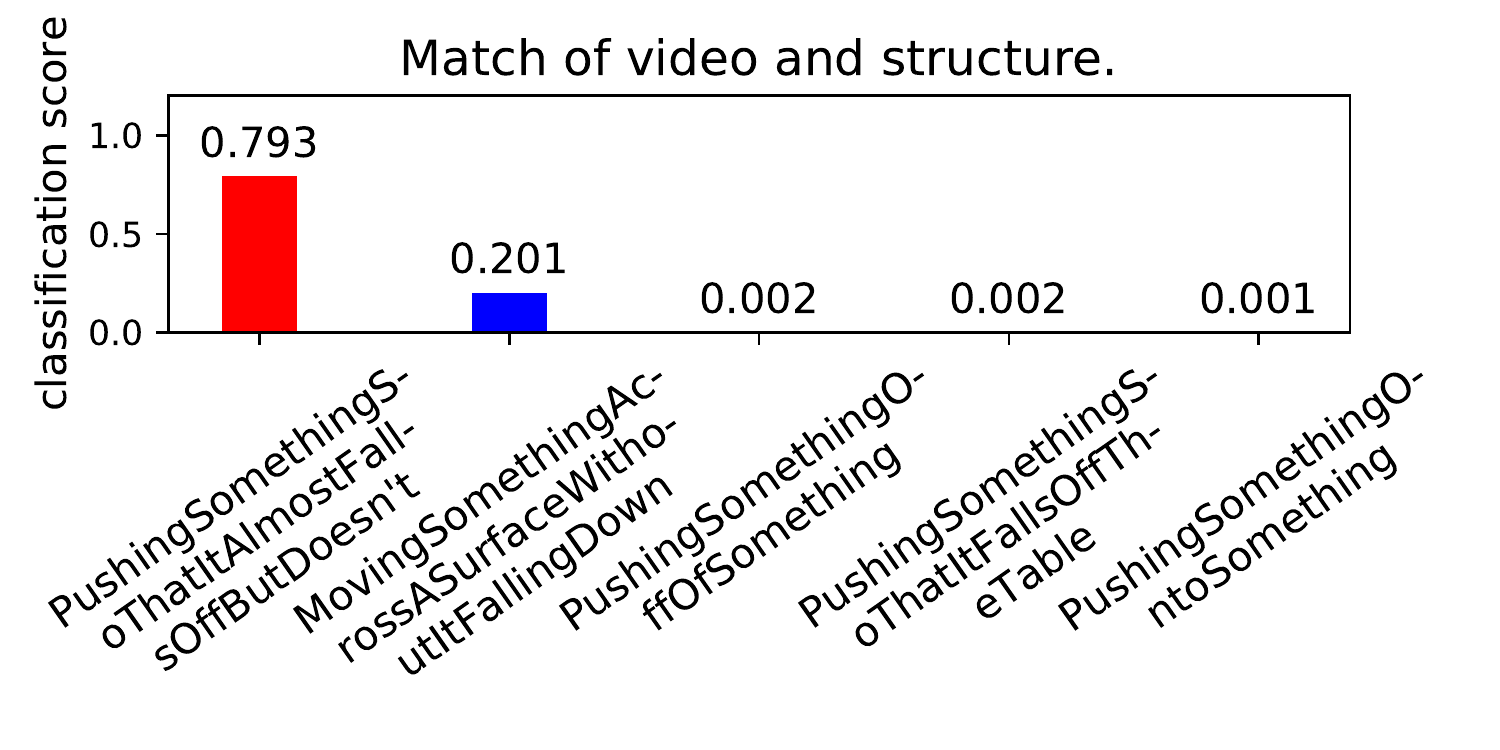}
    \includegraphics[width=0.47\linewidth]{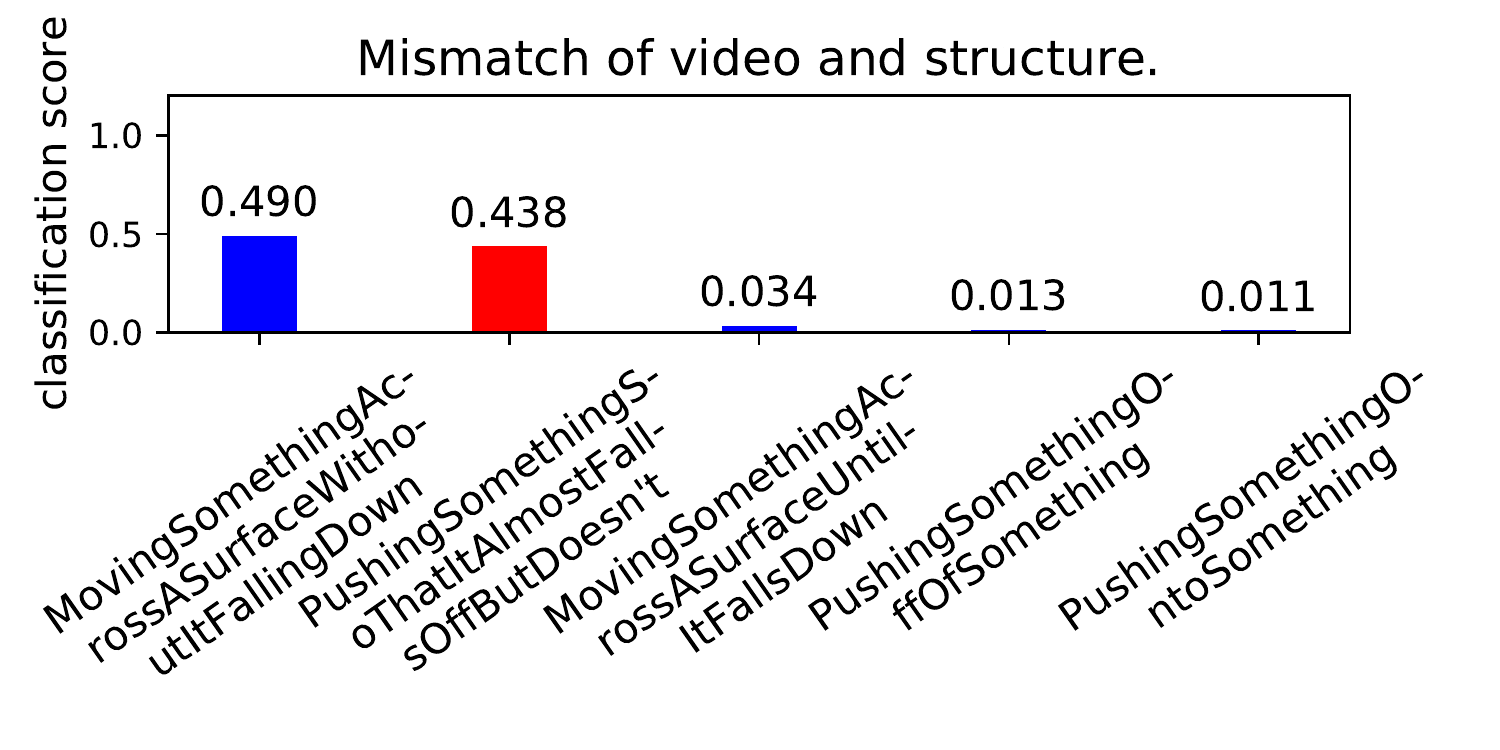}
    }

    \subfigure[Match and mismatch classification comparison of case 2.]{
    \includegraphics[width=0.47\linewidth]{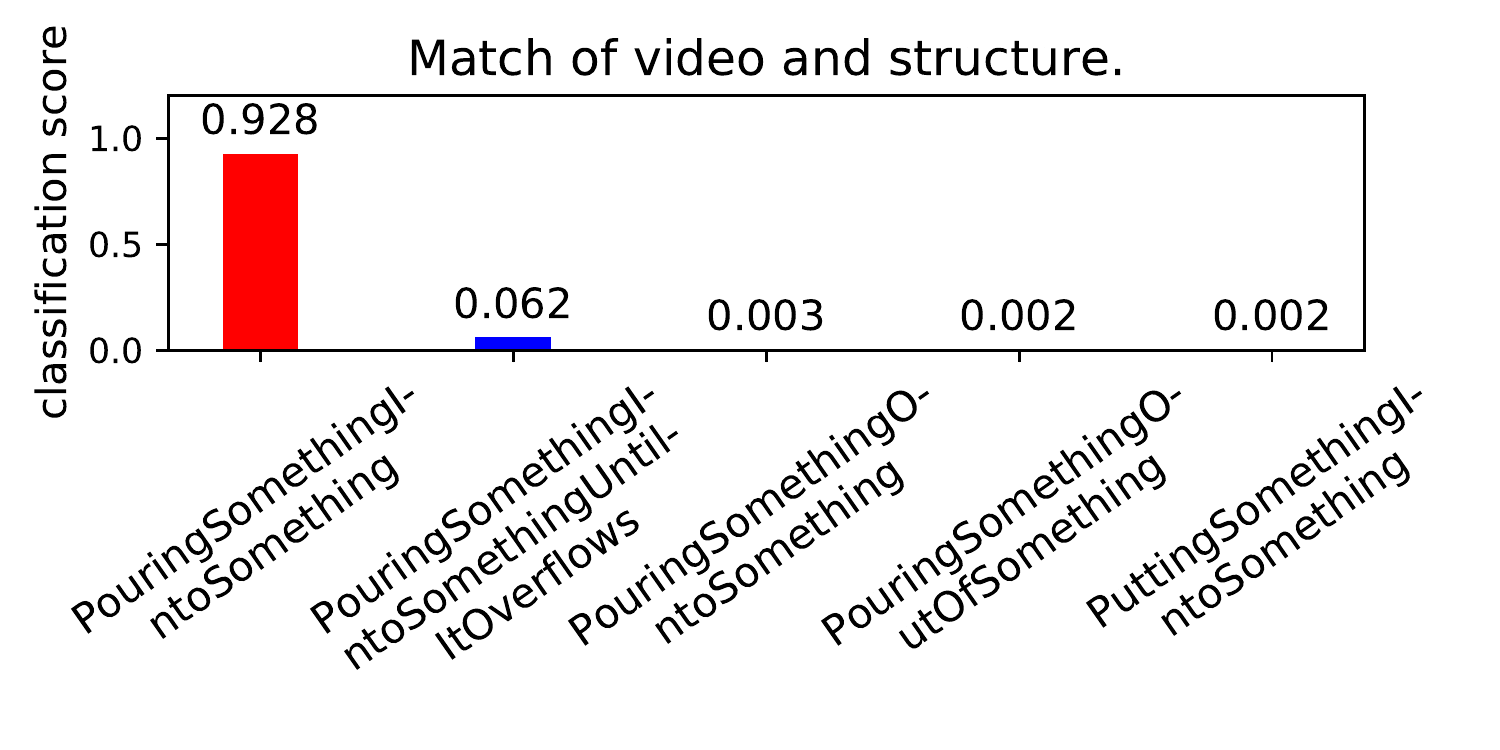}
    \includegraphics[width=0.47\linewidth]{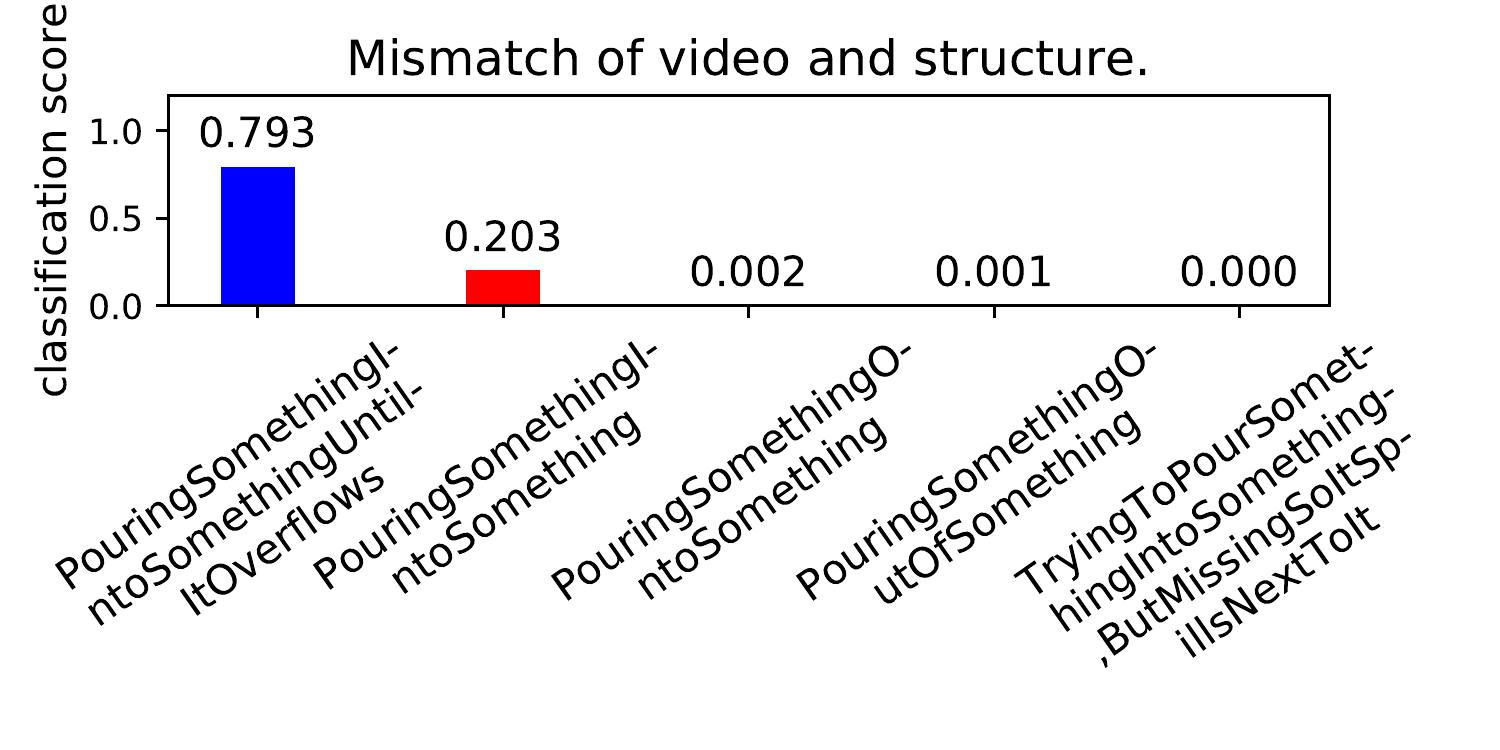}
    }
    \caption{Top 5 classification score comparison of match and mismatch of videos and structures. (a) and (b) show the results of the two cases in Figure \ref{fig:nas-ada-structure}. The red bars indicate the groundtruth categories.}
    \label{fig:nas-ada-mismatch}
\end{figure}

\noindent
{\bf{Mismatch of videos and structures.}} To validate the specificity of adaptive structures, we swap the two searched structures in Figure \ref{fig:nas-ada-structure} to mismatch the input videos, and use them to recognize the interactions. The results are compared in Figure \ref{fig:nas-ada-mismatch}. We observe that the mismatch of videos and structures leads to misclassification, which reveals that different videos require different structures for effective interaction modeling, since different interactions of different complexities are involved.

\begin{figure}[t]
    \centering
    \subfigure[Something-Something-V1]{
    \includegraphics[width=0.47\linewidth]{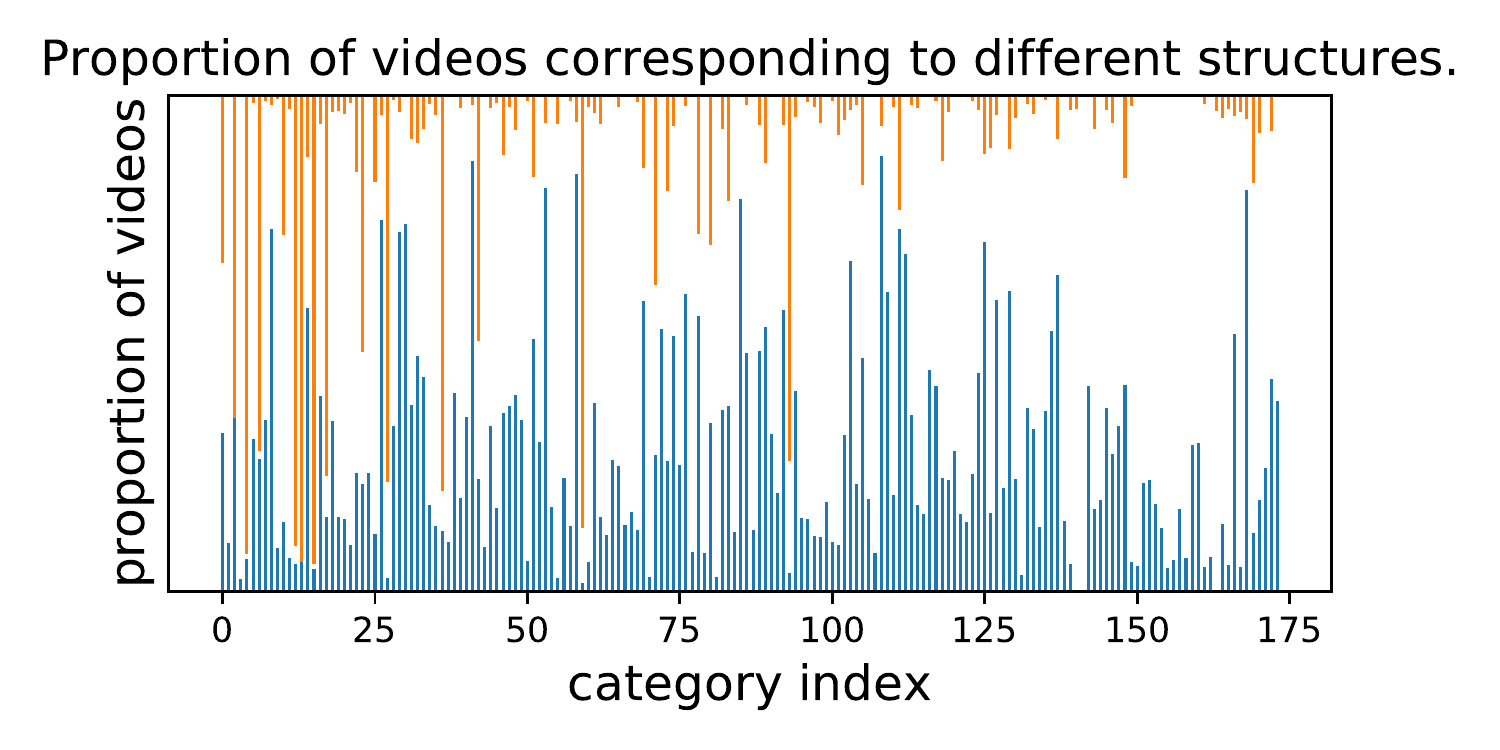}
    }
    \subfigure[Something-Something-V2]{
    \includegraphics[width=0.47\linewidth]{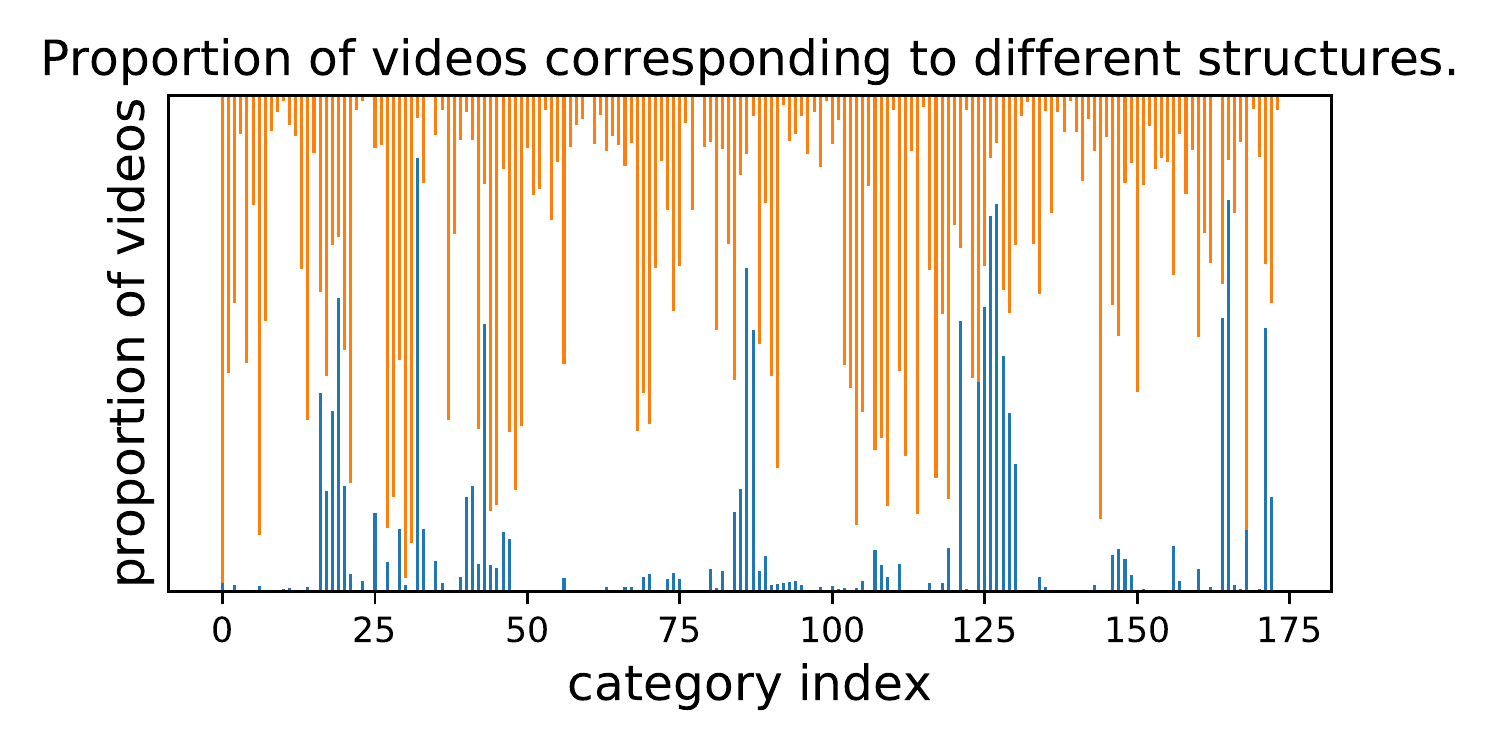}
    }
    \caption{The proportion of videos per class corresponding to different structures. (a) and (b) show the results on the two datasets. The bars with different colors indicate different structures.}
    \label{fig:nas-ada-statistic}
\end{figure}

\subsubsection{Analysis of Adaptive Structures}
To understand the relations between the adaptive structures and the interaction categories, we statistically analyze the proportion of videos per class corresponding to different searched structures in validation set. Figure \ref{fig:nas-ada-statistic} compares the results of two searched structures indicated with different colors. We observe that the searched structures are strongly correlated to the interaction categories, where each structure corresponds to  some specific interaction categories. For examples, in Something-Something-V1 dataset, the structure indicated with orange bars mainly corresponds to the interactions of indexes $\{2,4,6,12,15,\etal\}$, which are about the motions of the camera. While the structure indicated with blue bars includes the interactions about moving/pushing objects (of indexes $\{8,26,29,30,41,\etal\}$). This reveals that our architecture search framework learns to roughly divide the videos into several groups according to some characteristics in the interactions, and search specialized structures for different groups for adaptive interaction modeling. In other words, the adaptive structures automatically model interactions in a coarse (groups) to fine (specialized structure for each group) manner.

We further quantitatively compare the interaction recognition accuracy of non-adaptive and adaptive search schemes in Table \ref{tab:nas-scheme-cmp}. We make the following observations: On the one hand, adaptive scheme gains better performance than non-adaptive schemes. On the other hand, using only one searched structure for testing leads to obvious performance degradation, since different structures are searched to match different groups during training but only one structure is used for testing, which is insufficient to model the interactions in all groups. These observations further indicate the effectiveness of the adaptive structures.

We also validate that learning with fixed substructures gains slight improvements, diversity regularization helps to learn structures with multiple operations, and the adaptive structures can transfer across datasets. For more details, please refer to our supplementary materials.

\subsection{Analysis of Graph Operations}
In this section, we analyze the role of each graph operation in interaction modeling. Firstly, we compare the recognition accuracy of different operations by placing them on top of the backbone, and the results are shown in Table \ref{tab:cmp-operations}. It is seen that all the operations improve the performance over baselines, indicating that explicitly modeling the relations with graph operations benefits interaction recognition. Different graph operations gain different improvements, which depends on the significance of different relations in the datasets. In the following, we visualize some nodes and cases to demonstrate the different effects of different graph operations in interaction modeling.

\begin{table}[t]
    \scriptsize
    \centering
    \begin{tabular}{c|c|c}
        \hline
        Operations & V1 Val Acc & V2 Val Acc \\
        \hline\hline
        global pooling & 48.1 & 60.3 \\
        pooling over RoIs & 48.3 & 60.3 \\
        \hline
        feature aggregation & 49.9 & 62.0 \\
        difference propagation & 49.5 & 61.8 \\
        temporal convolution & 48.7 & 61.0 \\
        background incorporation & 49.7 & 62.4 \\
        node attention & 49.8 & 61.8 \\
        \hline
    \end{tabular}
    \caption{Interaction recognition accuracy (\%) comparison of different graph operations.}
    \label{tab:cmp-operations}
\end{table}

\begin{figure}[t]
    \scriptsize
    \centering
    \subfigure[Feature Aggregation]{
    \label{fig:nodes-vis-feat}
    \includegraphics[width=0.3\linewidth,height=0.14\linewidth]{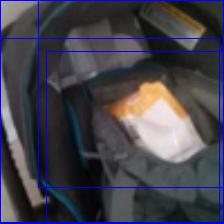}
    \includegraphics[width=0.3\linewidth,height=0.14\linewidth]{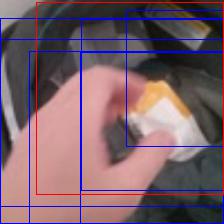}
    \includegraphics[width=0.3\linewidth,height=0.14\linewidth]{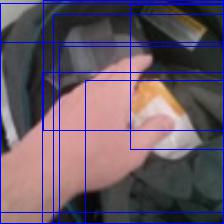}
    }
    \subfigure[Different Propagation]{
    \label{fig:nodes-vis-diff}
    \includegraphics[width=0.3\linewidth,height=0.14\linewidth]{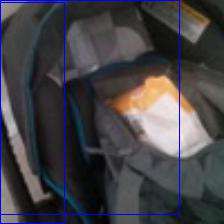}
    \includegraphics[width=0.3\linewidth,height=0.14\linewidth]{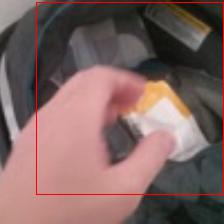}
    \includegraphics[width=0.3\linewidth,height=0.14\linewidth]{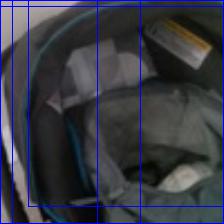}
    }
    \subfigure[Node Attention]{
    \label{fig:nodes-vis-att}
    \includegraphics[width=0.3\linewidth,height=0.14\linewidth]{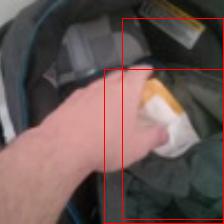}
    \includegraphics[width=0.3\linewidth,height=0.14\linewidth]{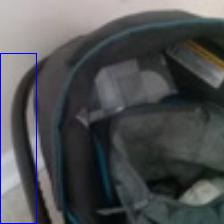}
    \includegraphics[width=0.3\linewidth,height=0.14\linewidth]{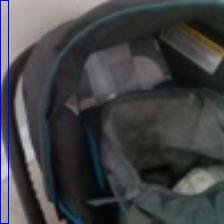}
    }
    \caption{\footnotesize{Top activated nodes of different operations on the same interaction ``Pulling something out of something''. In (a) and (b), the red node is the reference node and the blue nodes are the top activated nodes. In (c), The red nodes have the highest attention weights while the blue ones have the lowest attention weights.}}
    \label{fig:nodes-vis}
\end{figure}

\noindent
{\bf{Top activated nodes.}} We visualize the nodes with top affinity values of some operations for the same video in Figure \ref{fig:nodes-vis}. The \textit{feature aggregation} focuses on the apparently similar nodes to model the dependencies among them as shown in Figure \ref{fig:nodes-vis-feat}. On the contrary, the \textit{difference propagation} \revise{models the significant changes of some obviously different nodes in Figure \ref{fig:nodes-vis-diff}. In Figure \ref{fig:nodes-vis-att}, the nodes with high attention weights are the hand or the bag, and the nodes with low attention weights are some outliers, which indicates that the \textit{node attention} helps to concentrate on important nodes and eliminate the interference of outliers.}

\begin{figure}[t]
    \scriptsize
    \centering
    \subfigure[Stuffing something into something]{
    \label{fig:case-study-gcn}
    \includegraphics[width=0.24\linewidth]{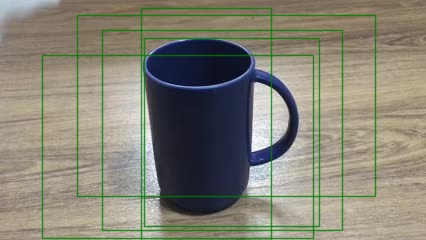}
    \includegraphics[width=0.24\linewidth]{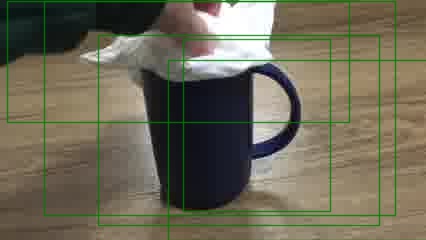}
    \includegraphics[width=0.24\linewidth]{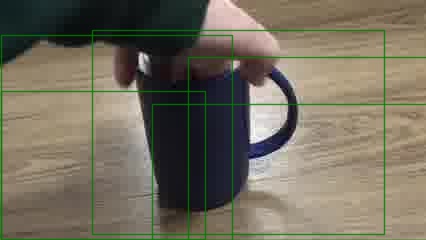}
    \includegraphics[width=0.24\linewidth]{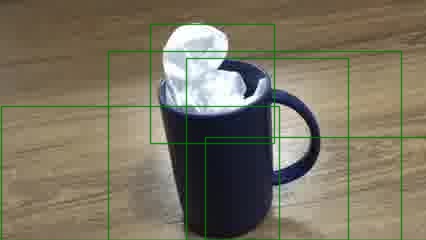}
    }
    \subfigure[Twisting something]{
    \label{fig:case-study-diff-prop}
    \includegraphics[width=0.24\linewidth]{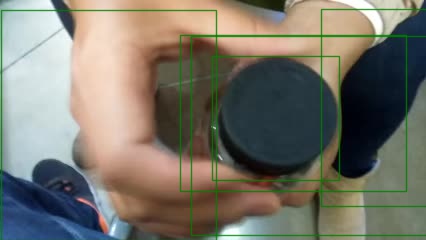}
    \includegraphics[width=0.24\linewidth]{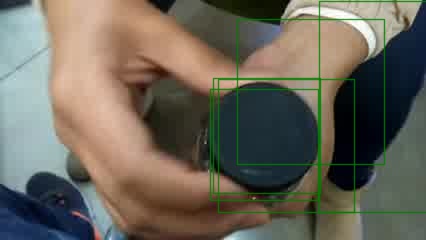}
    \includegraphics[width=0.24\linewidth]{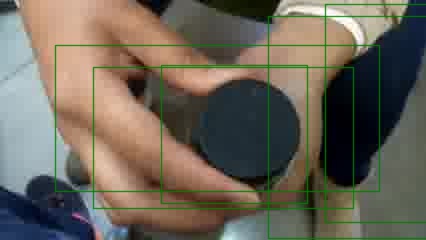}
    \includegraphics[width=0.24\linewidth]{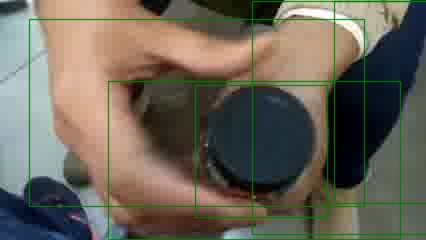}
    }
    \subfigure[Twisting something wet until water comes out]{
    \label{fig:case-study-back-agg}
    \includegraphics[width=0.24\linewidth]{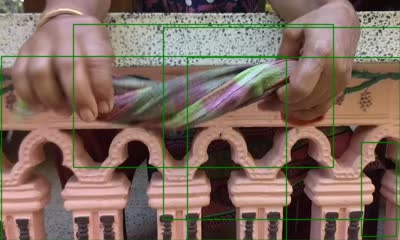}
    \includegraphics[width=0.24\linewidth]{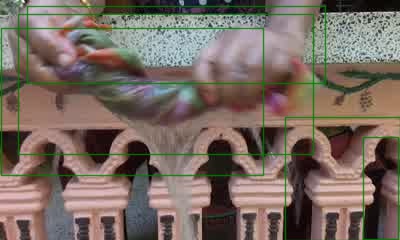}
    \includegraphics[width=0.24\linewidth]{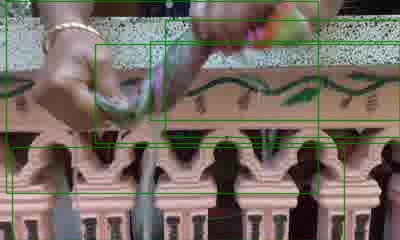}
    \includegraphics[width=0.24\linewidth]{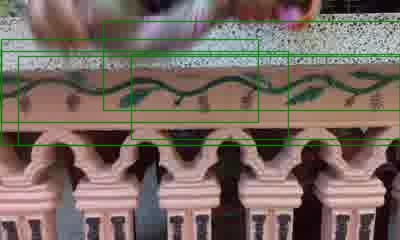}
    }
    \caption{\footnotesize{Successful and failed cases of different graph operations. The green bounding boxes are RoIs extracted from RPN.}}
    \label{fig:case-study}
\end{figure}

\noindent
{\bf{Successful and failed cases.}} We show some successful and failed cases to indicate the effects of different operations in Figure \ref{fig:case-study}. In Figure \ref{fig:case-study-gcn}, the \textit{feature aggregation} successfully recognizes the interaction \revise{due to the obvious dependencies between the paper and the mug}. However, it fails \revise{when detailed relations in Figure \ref{fig:case-study-diff-prop} and \ref{fig:case-study-back-agg} are present}. In Figure \ref{fig:case-study-diff-prop}, the \textit{difference propagation} and the \textit{temporal convolution} could capture that the lid is rotating so that they correctly recognize the interaction. In Figure \ref{fig:case-study-back-agg}, the \textit{background incorporation} is able to capture the relations between the towel and the water in the background so that it makes correct prediction, but other operations ignoring the background information are hard to recognize such an interaction with the background.

More case study and analysis about graph operations are included in supplementary materials.

\begin{table}[t]
    \footnotesize
    \centering
    \begin{threeparttable}
    \begin{tabular}{c|c|c}
        \hline
        Methods & V1 Val Acc & V2 Val Acc \\
        \hline\hline
        I3D+GCN \cite{Wang_2018_ECCV} (ECCV'18) & 43.3 & - \\
        NonLocalI3D+GCN \cite{Wang_2018_ECCV} (ECCV'18) & 46.1 & - \\
        CPNet \cite{Liu_2019_CVPR} (CVPR'19) & - & 57.6 \\
        TSM \cite{lin2018temporal} (ICCV'19) & 44.8\tnote{1} & 58.7\tnote{1} \\
        ECO \cite{Zolfaghari_2018_ECCV} (ECCV'18) & 46.4 & - \\
        TrajectoryNet \cite{NIPS2018_7489} (NeurIPS'18) & 47.8 & - \\
        S3D \cite{Xie_2018_ECCV} (ECCV'18) & 48.2 & - \\
        ir-CSN-152 \cite{tran2019video} (ICCV'19) & 48.4 & - \\
        GST \cite{Luo_2019_ICCV} (ICCV'19) & 48.6 & 62.6 \\
        discriminative filters \cite{martinez2019action} (ICCV'19) & 50.1\tnote{2} & - \\
        STM \cite{jiang2019stm} (ICCV'19) & 50.7 & \bf{64.2} \\
        adaptive structures search (Ours) & \bf{51.4} & 63.5 \\
        \hline
    \end{tabular}
    \begin{tablenotes}
    \scriptsize
     \item[1] Only RGB results are reported for fair comparison.
     \item[2] Only the results with the same backbone (ResNet50) as ours are reported.
   \end{tablenotes}
   \end{threeparttable}
    \caption{Interaction recognition accuracy (\%) comparison with state-of-the-arts.}
    \label{tab:cmp-sota}
\end{table}

\subsection{Comparison with State-of-the-arts}
We compare the interaction recognition accuracy with recent state-of-the-art methods, and the results are show in Table \ref{tab:cmp-sota}. Except for STM \cite{jiang2019stm}, our method outperforms other methods, which indicates the effectiveness of our method. We model the interactions with adaptive structures, which enhances the ability of interaction modeling and boosts the performance.

Among the recent state-of-the-arts, I3D+GCN \cite{Wang_2018_ECCV} also uses graph operation over object proposals to recognize interactions. Our method surpasses it with a margin about 7\%, perhaps because we have trained a better backbone with our data augmentation techniques (see Section \ref{subsec:impl-details} for details), and our adaptive structures with multiple graph operations learn better interaction representations.

STM \cite{jiang2019stm} proposes a block to encode spatiotemporal and motion features, and stacks it into a deep network, which obtains better performance on Something-something-V2 dataset than ours. However, we adaptively model interactions with different structures, which provides more understanding about the relations between the interactions and the corresponding structures, instead of only feature encoding in STM. In addition, our structures are automatically searched, which releases the structures design efforts.


\section{Conclusion}

In this paper, we propose to automatically search adaptive network structures for interaction recognition, which enables adaptive interaction modeling and reduces structures design efforts. We design the search space with several proposed graph operations, and employ differentiable architecture search mechanism to search adaptive interaction modeling structures. Our experiments show that the architecture search framework learns adaptive structures for different videos, helping us understand the relations between structures and interactions. In addition, the designed basic graph operations model different relations in videos. The searched adaptive structures obtain competitive interaction recognition performance with state-of-the-arts.

\section*{Acknowledgement}

This work was supported partially by the National Key Research and Development Program of China (2018YFB1004903), NSFC(U1911401,U1811461), Guangdong Province Science and Technology Innovation Leading Talents (2016TX03X157), Guangdong NSF Project (No. 2018B030312002), Guangzhou Research Project (201902010037), and Research Projects of Zhejiang Lab (No. 2019KD0AB03). The principal investigator for this work is Wei-Shi Zheng.


\clearpage

\setcounter{section}{0}
\setcounter{figure}{0}
\setcounter{table}{0}

\twocolumn[
    \begin{@twocolumnfalse}
    \begin{center}
        \Large \textbf{Supplementary Material for: \\
        Adaptive Interaction Modeling via Graph Operations Search \\}
        \vspace{1em}
    \end{center}
    \end{@twocolumnfalse}
]

%
%
%

\revise{
\section{Additional Results}
\subsection{Results on Test Set}
We report the results on validation set in our paper for comprehensive comparison, because most of other methods only report validation results due to the withheld test labels. In Table \ref{tab:test-result}, we compare the results on test set, which show that our method also achieves competitive performances with other available methods.
\begin{table}[h]
\centering
\begin{tabular}{ccc}
\hline
Methods&V1 test acc&V2 test acc\\
\hline
NonLocalI3D+GCN&45.0&- \\
CPNet&-&57.6 \\
TSM&46.1&59.9 \\
ECO&42.3&- \\
S3D&42.0&- \\
GST&-&61.2 \\
STM&43.1&63.5 \\
Ours&46.3&62.6 \\
\hline
\end{tabular}
\caption{Top-1 accuracy (\%) of RGB based methods on test set.}
\label{tab:test-result}
\end{table}

\subsection{Model Size and Inference Time}
In Table \ref{tab:FLOPs}, we evaluate the number of parameters and MACs (multiply–accumulate operations) of our model using a public available tool (\url{https://github.com/sovrasov/flops-counter.pytorch}). Our method would not increase the model size too much (comparable with NonLocalI3D), but still boost performance.
\begin{table}[h]
\centering
\begin{threeparttable}[b]
\begin{tabular}{ccc}
\hline
Model&Params&GMACs\\
\hline
NonLocalI3D&35.3 M\tnote{$\dagger$}&167.5\tnote{$\dagger$} \\
\hline
Our backbone&32.1 M&127.6 \\
Our whole model&37.3 M&138.4 \\
\hline
\end{tabular}
\caption{Number of parameters and MACs ([$\dagger$] Calculated in \cite{lin2018temporal}).}
\label{tab:FLOPs}
\end{threeparttable}
\end{table}

In terms of inference time, our framework takes around 0.12 seconds per video with 32 sampled frames on a single GTX 1080TI GPU, which is still towards real time and would not lead to excessive latency.
}

\section{Implementation Details}
\noindent
{\bf{Backbone.}} We use I3D-ResNet \cite{8578911,Wang_2018_ECCV}, which inflates 2D convolution kernels into 3D kernels for initialization \cite{8099985}, as our backbone (Table \ref{tab:backbone}) to extract basic features. It is inflated from ResNet-50 \cite{7780459} with ImageNet \cite{deng2009imagenet} pretrained parameters, and it extracts video features after ``res5" with 2048 channels from 32 uniformly sampled frames. For computation efficiency in graph reasoning, we reduce the feature dimension from 2048 to 256 with a FC layer.
\begin{table}[h]
\centering
\begin{tabular}{c|c|c}
    \hline
    \multicolumn{2}{c|}{layer} & output size \\
    \hline\hline
    conv1 & 5$\times$7$\times$7, 64, stride 1,2,2 & 32$\times$112$\times$112 \\
    \hline
    pool1 & 3$\times$3$\times$3, max, stride 1,2,2 & 32$\times$56$\times$56 \\
    \hline
    res2 &
    $\left[
    \begin{array}{c}
    3\times 1\times 1, 64 \\
    1\times 3\times 3, 64 \\
    1\times 1\times 1, 256
    \end{array}
    \right]\times 3$ & 32$\times$56$\times$56 \\
    \hline
    pool2 & 3$\times$1$\times$1, max, stride 2,1,1 & 16$\times$56$\times$56 \\
    \hline
    res3 &
    $\left[
    \begin{array}{c}
    3\times 1\times 1, 128 \\
    1\times 3\times 3, 128 \\
    1\times 1\times 1, 512
    \end{array}
    \right]\times 4$ & 16$\times$28$\times$28 \\
    \hline
    res4 &
    $\left[
    \begin{array}{c}
    3\times 1\times 1, 256 \\
    1\times 3\times 3, 256 \\
    1\times 1\times 1, 1024
    \end{array}
    \right]\times 6$ & 16$\times$14$\times$14 \\
    \hline
    res5 &
    $\left[
    \begin{array}{c}
    3\times 1\times 1, 512 \\
    1\times 3\times 3, 512 \\
    1\times 1\times 1, 2048
    \end{array}
    \right]\times 3$ & 16$\times$7$\times$7 \\
    \hline
\end{tabular}
\caption{Our backbone ResNet-50 I3D model. The kernel size and output size are represented as $T\times H \times W$. The input size is 32$\times$224$\times$224.}
\label{tab:backbone}
\end{table}

\noindent
{\bf{RPN.}} We use RPN \cite{He_2017_ICCV} model with ResNet-50 and FPN to extract region proposals. The RPN model is pre-trained on the MSCOCO object detection dataset \cite{lin2014microsoft}. To match the output time dimension of the ``res5'' feature maps, we sample 16 frames from 32 input frames (1 frame every 2 frames) to extract region proposals. Top $10$ class-agnostic object bounding boxes are extracted for each frame.

\noindent
{\bf{Graph operations settings.}}
The shapes of all the transform matrices $\boldsymbol{W}_{*}$, $\boldsymbol{V}_{*}$ in the graph operations are set as $256\times C_{in}$, where $C_{in}$ differs from operation to operation. The shapes of the affinity weights $\boldsymbol{U}_{*}$ are set as $C_{in}\times C_{in}$. The affinity matrices are row-normalized to keep the sum of the affinities connected to each node to be 1. The size of temporal convolution kernel $\boldsymbol{W}_t$ is set to 7. We employ Layer Normalization \cite{ba2016layer} followed by LeakyReLU as the nonlinear activation function of each operation.

\noindent
{\bf{Training.}}
We train our model in the following steps:
\begin{enumerate}
\setlength{\itemsep}{0pt}
\setlength{\parsep}{0pt}
\setlength{\parskip}{0pt}
  \item Train the backbone on the target datasets.
  \item Fix the backbone. Train the weights in all the graph operations and the structure weights alternatively to learn adaptive structures until the adaptive structures are stable. The SGD optimizer is used for the weights in all the graph operations, and the Adam optimizer is used for the structure weights. The learning rate of the weights in all the graph operations is 0.01, and the learning rate of the structure weights is 0.0001.
  \item Fix the structure weights. Train the weights in all the graph operations with discrete structures. SGD optimizer is used and the learning rate is set to 0.001. The learning rate is divided by 10 when the validation loss doesn't decline for 5 epochs. The training is stopped when the validation loss doesn't decline for 5 epochs with learning rate 0.0001.
  \item Train the weights in all the graph operations and the backbone jointly. SGD optimizer is used and the learning rate is 0.0001.
\end{enumerate}

\noindent
{\bf{Testing.}}
On the testing stage, we sample 5 clips in each video and use the mean score for classification.

\revise{
\noindent
{\bf{Data augmentation.}}We divide the video into 32 segments and randomly sample one frame in each segment, in order to obtain different samples from the same video for augmentation. We also randomly crop and horizontally flip all the sampled frames of the same video. It should be noted that some categories are relevant to directions, so that we do not apply horizontal flipping to these videos.
}

\section{Analysis of Architecture Search Framework}
\subsection{Fixing Substructures}
Table \ref{tab:nas-fixing-cmp} compares the performance of learning with original graph operations and fixed substructures. It is observed that learning with fixed substructures obtains higher accuracy, perhaps because it simplifies the optimization with fewer structure weights and also implicitly deepens the structures. In addition, learning with fixed substructures converges faster than learning with original graph operations in our experiments, which reduces the searching time.
\begin{table}[h]
    \centering
    \begin{tabular}{c|c|c}
        \hline
        Settings & V1 Val Acc & V2 Val Acc \\
        \hline\hline
        Ori Ops & 51.0 & 63.1 \\
        Fixed Subs & 51.4 & 63.5 \\
        \hline
    \end{tabular}
    \caption{Interaction recognition accuracy (\%) comparison of different search space. ``Ori Ops'' means original graph operations are used as basic operations in the search space, and ``Fixed Subs'' means the fixed substructures are used as basic operations to search the structures.}
    \label{tab:nas-fixing-cmp}
\end{table}

\subsection{Diversity Regularization}
\begin{figure}[h]
    \centering
    \includegraphics[width=\linewidth]{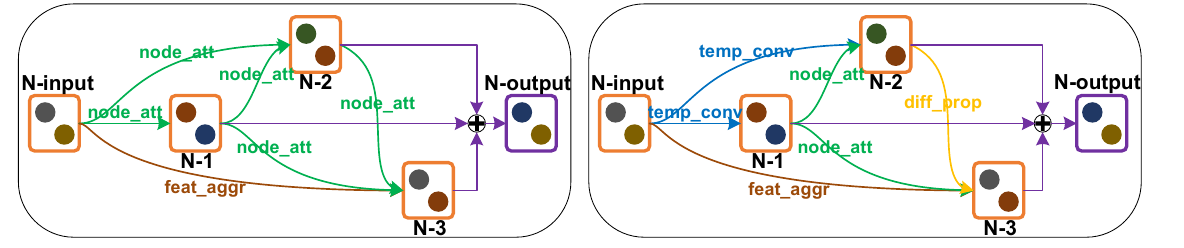}
    \caption{The learned structures without (left) and with (right) variance loss for diversity regularization.}
    \label{fig:nas-loss-cmp}
\end{figure}

To validate the effect of diversity regularization, we search non-adaptive structures with original graph operations on Something-Something-V1 dataset and compare the searched structures without and with diversity regularization (variance loss in Equation (13) in our paper) in Figure \ref{fig:nas-loss-cmp}. It is observed that the structure learned without variance loss tends to only select ``node attention", which hampers complex relation modeling and also obtains unsatisfactory performance (Table \ref{tab:nas-loss-cmp}). In contrast, the structure learned with variance loss selects diverse graph operations, which enhances the ability of interaction modeling and gains better recognition results. Therefore, we all use diversity regularization in other experiments.

\begin{table}[h]
    \centering
    \begin{tabular}{c|c}
        \hline
        Settings & V1 Val Acc \\
        \hline\hline
        w/o var loss & 49.9  \\
        with var loss & 50.7  \\
        \hline
    \end{tabular}
    \caption{Interaction recognition accuracy (\%) comparison about diversity regularization.}
    \label{tab:nas-loss-cmp}
\end{table}

\subsection{Transferability of Adaptive Structures}
In order to verify the transferability of the adaptive structures, we learn the adaptive structures on one dataset, and then fix the structure weights and train the rest learnable weights on the other dataset. The results are shown in Table \ref{tab:nas-transfer-cmp}. It is observed that the interaction recognition performance does not decline obviously (0.4\% for Something-Something-V1 dataset and 0.1 \% for Something-Something-V2 dataset), which indicates that the adaptive structures can transfer across datasets with minor performance degradation.
\begin{table}[h]
    \centering
    \begin{tabular}{c|c|c}
        \hline
        transfer settings & V2 $\rightarrow$ V1 & V1 $\rightarrow$ V2 \\
        \hline\hline
        accuracy & 51.0 & 63.4 \\
        \hline
    \end{tabular}
    \caption{Interaction recognition accuracy (\%) of transferring the adaptive structures across datasets. ``dataset1 $\rightarrow$ dataset2'' means the structure weights are trained in dataset1 and fixed in dataset2.}
    \label{tab:nas-transfer-cmp}
\end{table}

\begin{figure}[t]
    \centering
    \subfigure[Structure\quad\quad\quad\quad\quad\quad\quad\quad\quad]{
    \label{fig:transfer-1-stru}
    \includegraphics[width=0.95\linewidth]{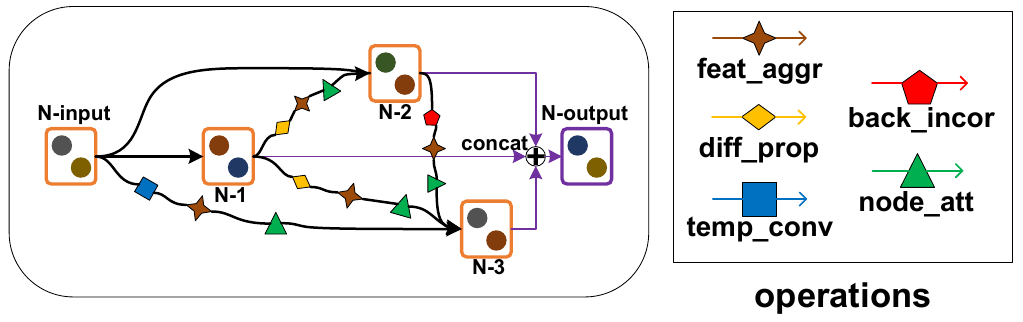}
    }
    \subfigure[Something-Something-V1]{
    \label{fig:transfer-1-ori}
    \includegraphics[width=0.47\linewidth]{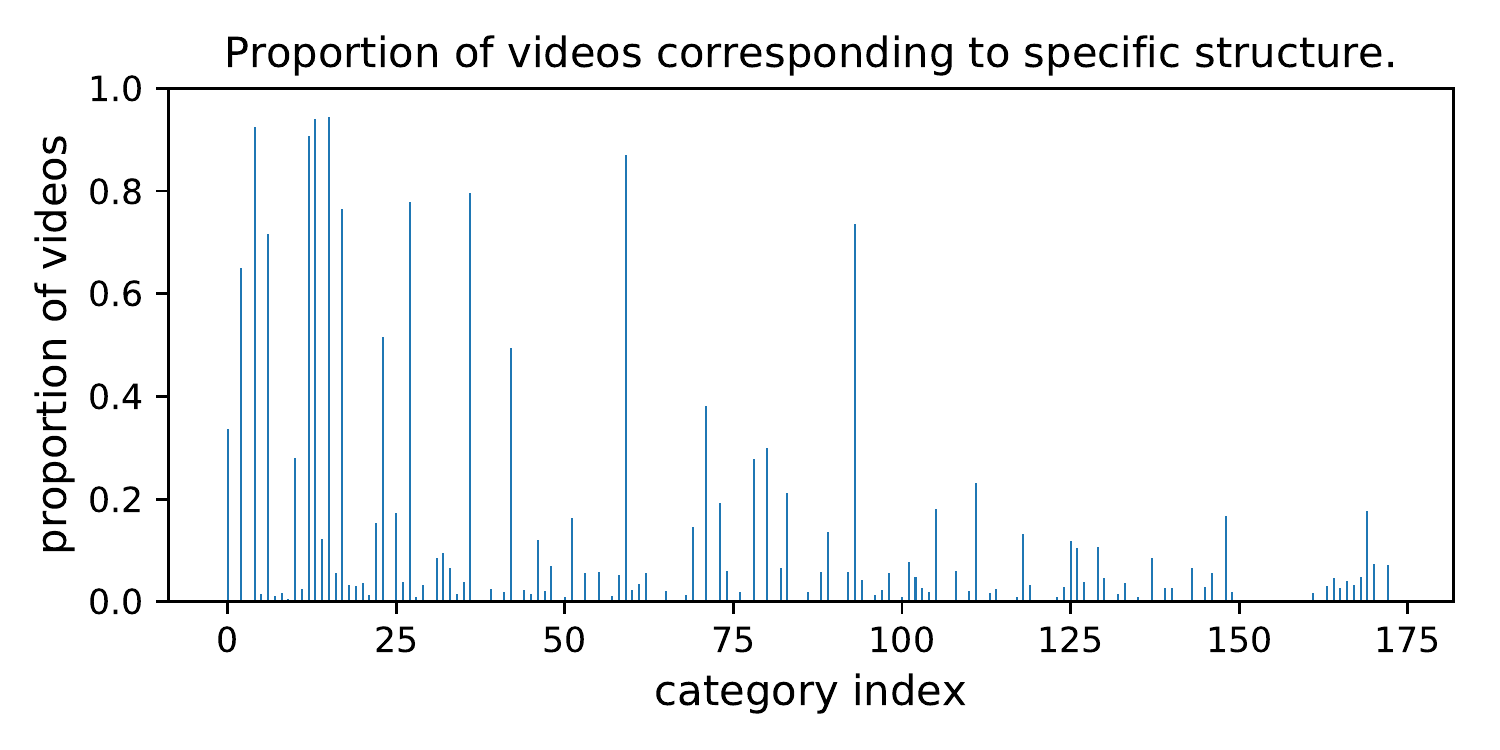}
    }
    \subfigure[Something-Something-V2]{
    \label{fig:transfer-1-tra}
    \includegraphics[width=0.47\linewidth]{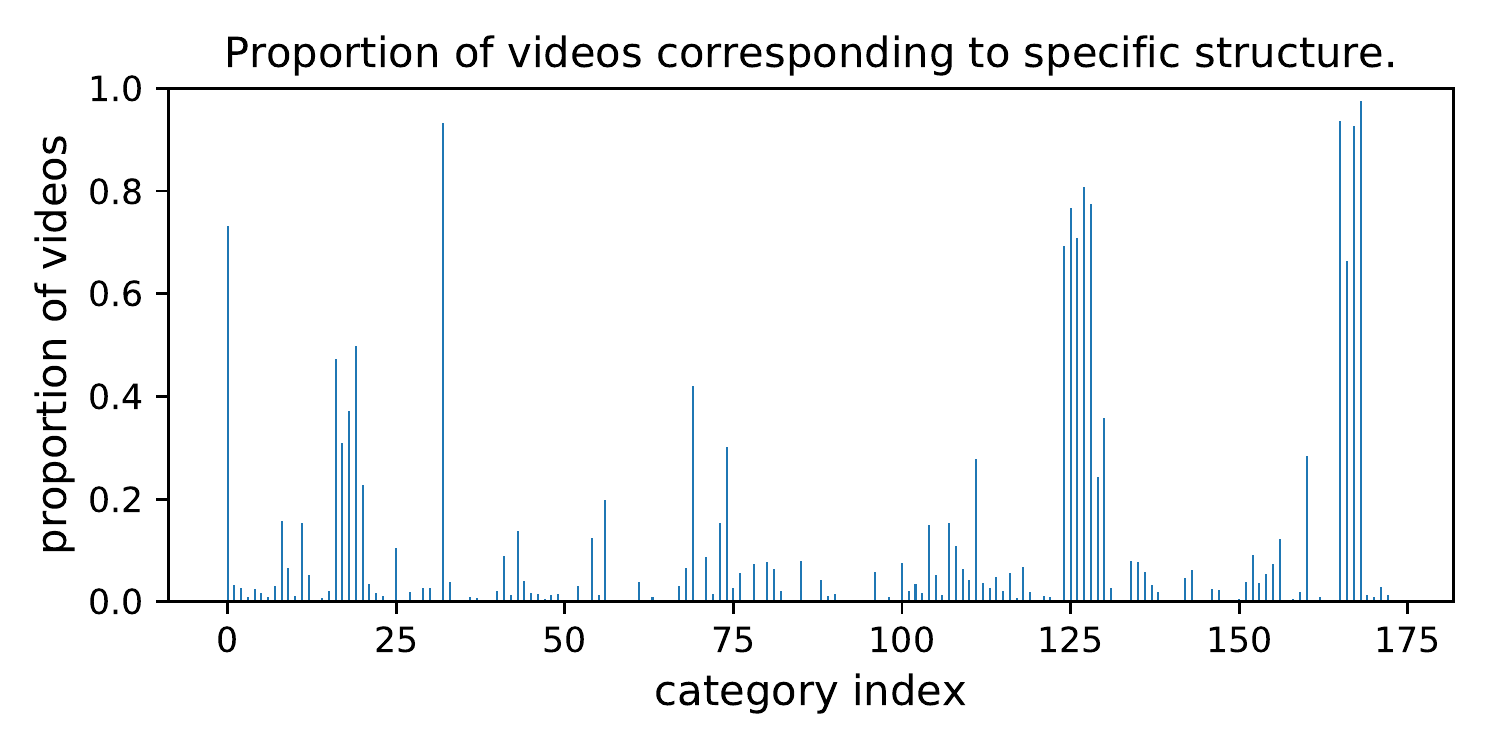}
    }
    \caption{Example 1: proportion of videos per class corresponding to the structure (a) in the original dataset (b) and the transferred dataset (c).}
    \label{fig:transfer-1}
\end{figure}

\begin{figure}[t]
    \centering
    \subfigure[Structure\quad\quad\quad\quad\quad\quad\quad\quad\quad]{
    \label{fig:transfer-2-stru}
    \includegraphics[width=0.95\linewidth]{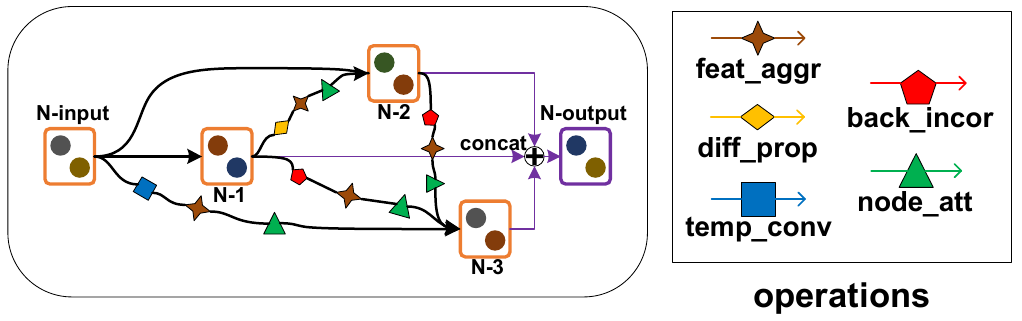}
    }
    \subfigure[Something-Something-V1]{
    \label{fig:transfer-2-ori}
    \includegraphics[width=0.47\linewidth]{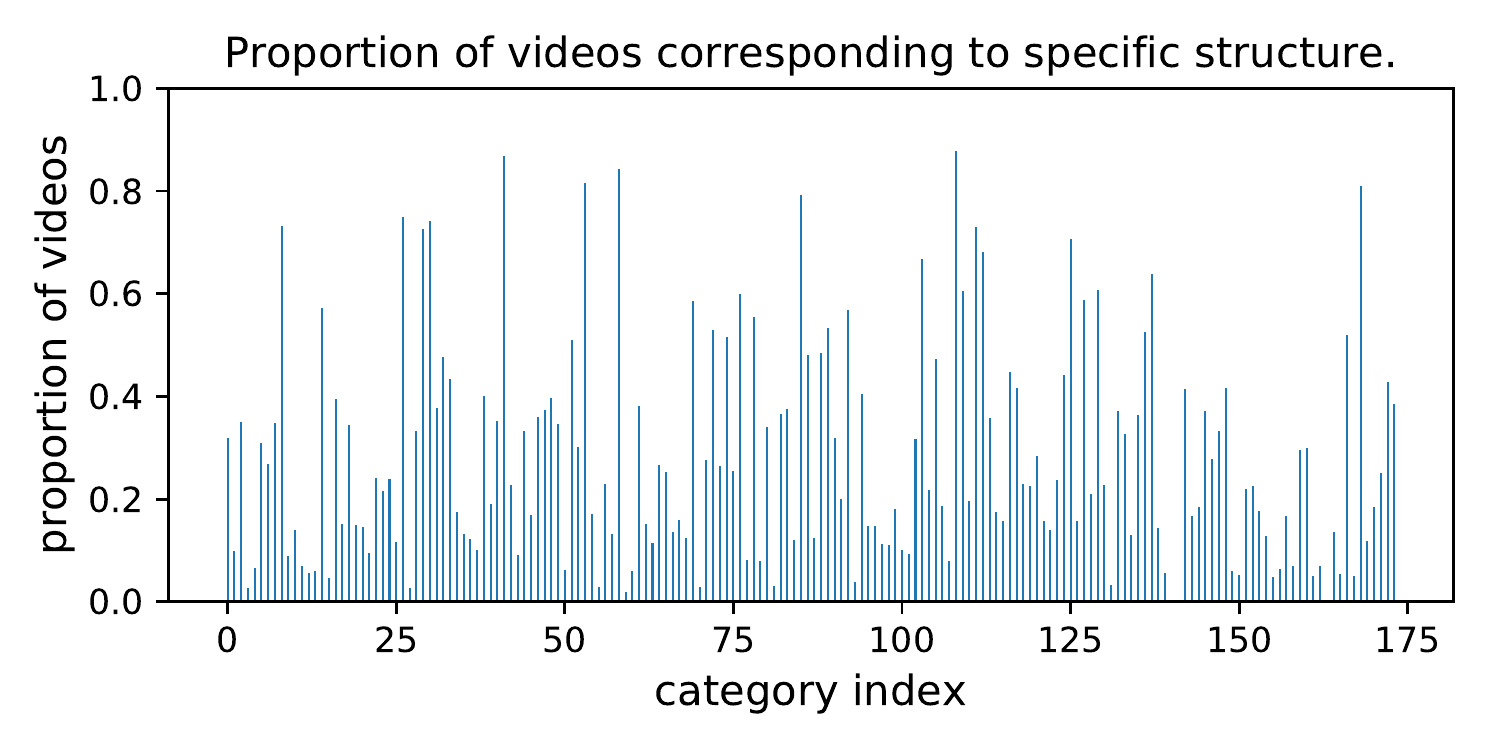}
    }
    \subfigure[Something-Something-V2]{
    \label{fig:transfer-2-tra}
    \includegraphics[width=0.47\linewidth]{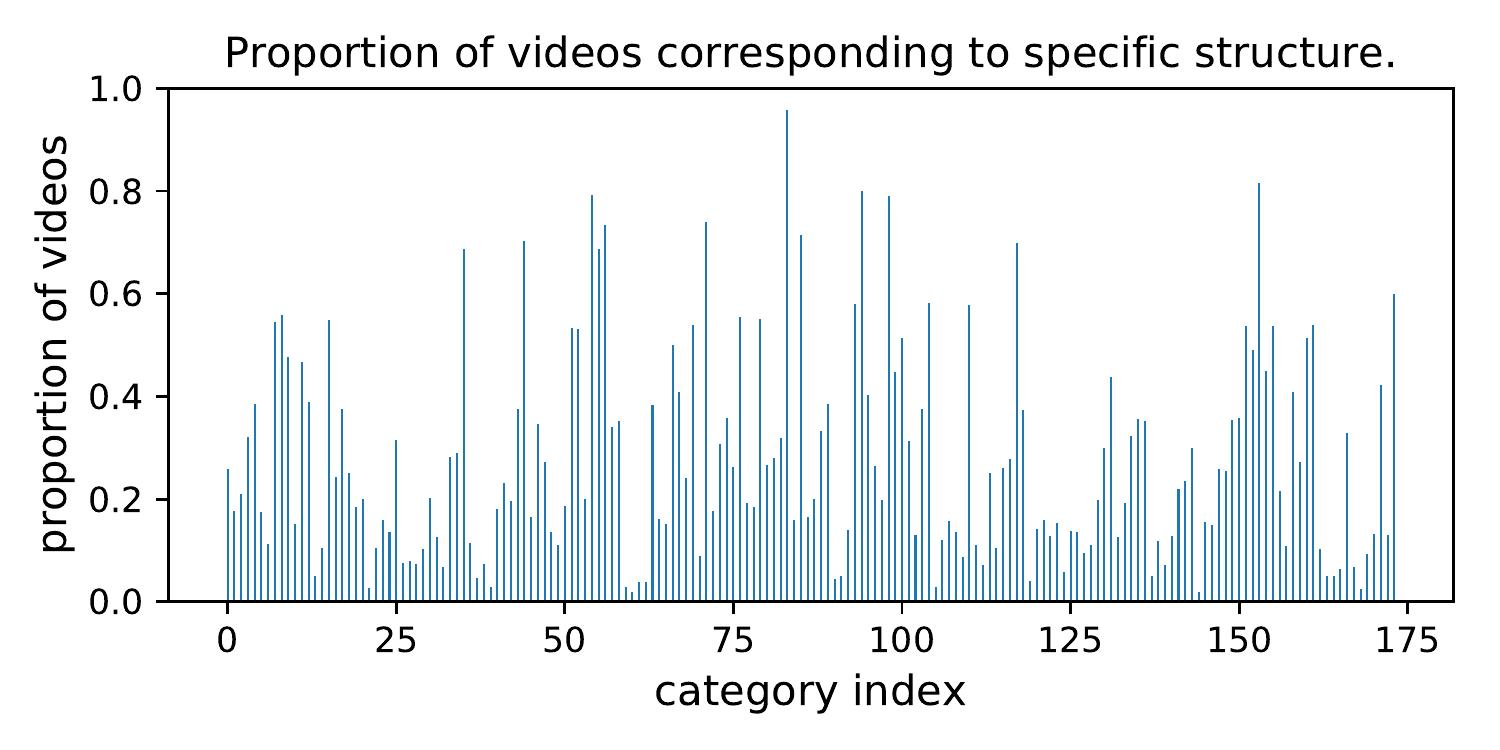}
    }
    \caption{Example 2: proportion of videos per class corresponding to the structure (a) in the original dataset (b) and the transferred dataset (c).}
    \label{fig:transfer-2}
\end{figure}

\begin{figure}[t]
    \centering
    \subfigure[Structure\quad\quad\quad\quad\quad\quad\quad\quad\quad]{
    \label{fig:transfer-3-stru}
    \includegraphics[width=0.95\linewidth]{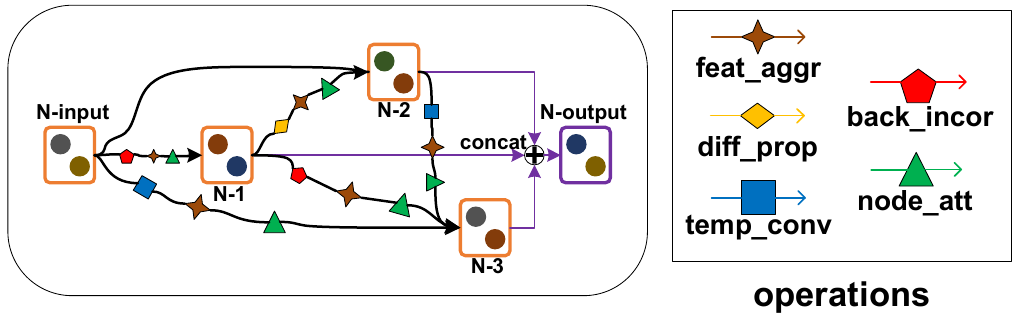}
    }
    \subfigure[Something-Something-V1]{
    \label{fig:transfer-3-ori}
    \includegraphics[width=0.47\linewidth]{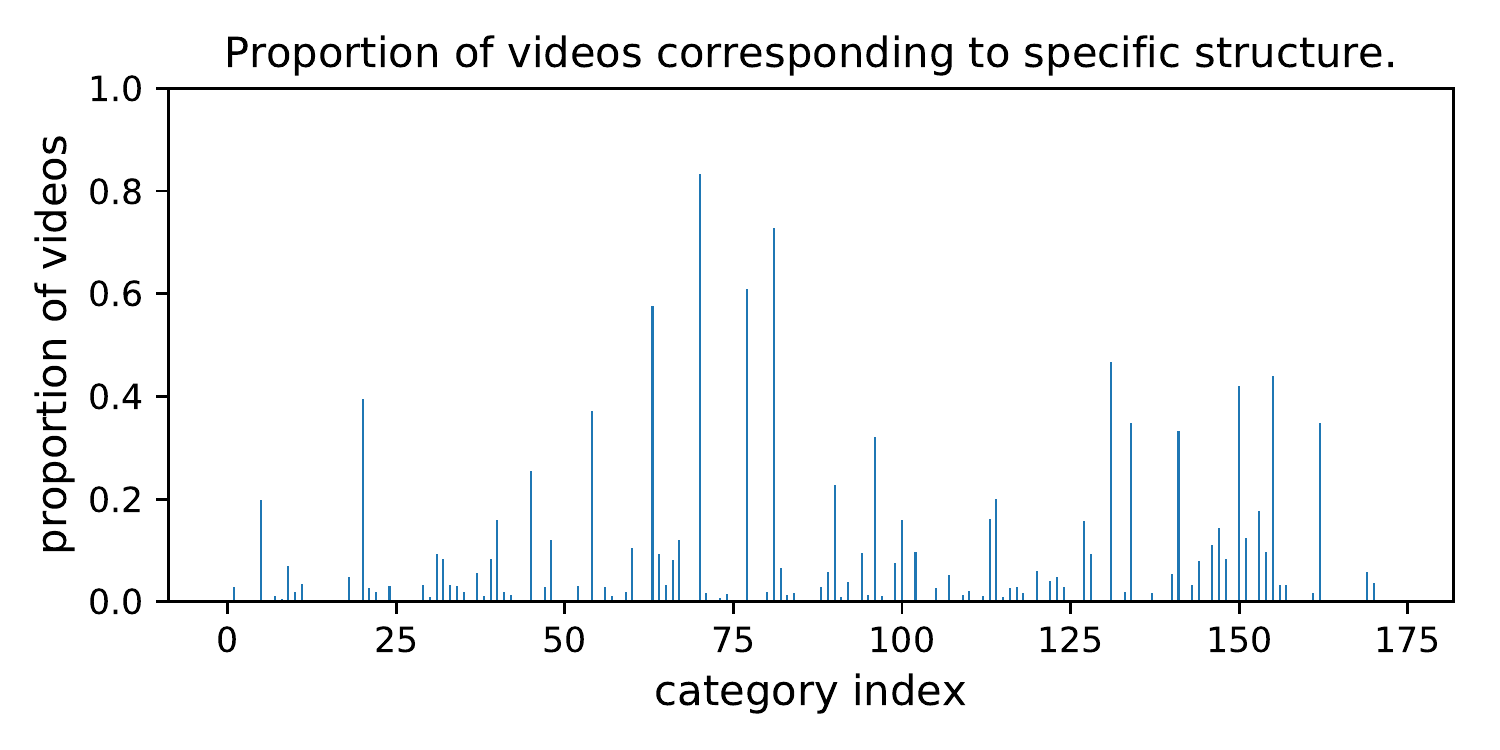}
    }
    \subfigure[Something-Something-V2]{
    \label{fig:transfer-3-tra}
    \includegraphics[width=0.47\linewidth]{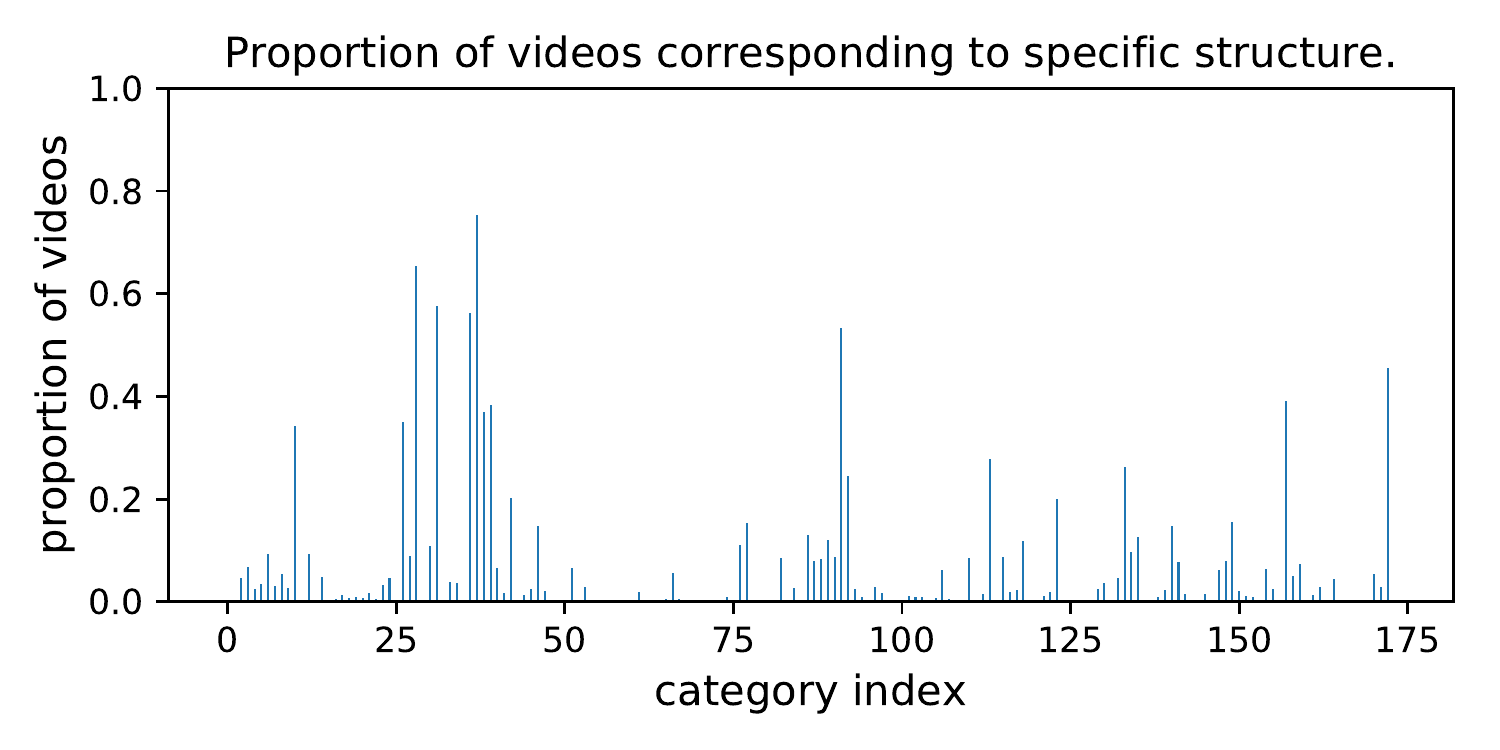}
    }
    \caption{Example 3: proportion of videos per class corresponding to the structure (a) in the original dataset (b) and the transferred dataset (c).}
    \label{fig:transfer-3}
\end{figure}

We further show the proportion of videos per class corresponding to some structures in the original and transferred datasets. Figure \ref{fig:transfer-1} to Figure \ref{fig:transfer-3} show three examples of the structure and its corresponding interaction category distributions in the original dataset (Something-Something-V1) and transferred dataset (Something-Something-V2). According to the category index and the label lists, we observed that the dominant interaction categories are semantically similar in the two datasets. In both the original dataset and the transferred dataset, the dominant categories are about camera motion, pushing/poking/throwing something, moving something away or closer in the three examples respectively. These results fully illustrate that the adaptive structures are learned according to some interaction characteristics, which can transfer across datasets and also help us understand the relations between the structures and the interactions.

\subsection{Computation Cell}
We compare the performance of the computation cells with different number of intermediate supernodes, and the results are shown in Table \ref{tab:nas-cell-cmp}. It is observed that computation cells with 3 and 4 intermediate supernodes obtain better performances than that with 2 intermediate supernodes. But using 4 intermediate supernodes could not further improve the performance, perhaps because the structures with 4 intermediate supernodes are too complex, which leads to highly complex back-propagation and optimization difficulties. Therefore, we use 3 intermediate supernodes in our experiments.
\begin{table}[h]
    \centering
    \begin{tabular}{c|c|c}
        \hline
        Settings & V1 Val Acc & V2 Val Acc \\
        \hline\hline
        2 supernodes & 50.6 & 63.1 \\
        3 supernodes & 51.4 & 63.5 \\
        4 supernodes & 51.3 & 63.5 \\
        \hline
    \end{tabular}
    \caption{Interaction recognition accuracy (\%) comparison of different number of intermediate supernodes in the computation cell.}
    \label{tab:nas-cell-cmp}
\end{table}

\subsection{Deeper Structures}
We attempt to stack multiple searched computation cells to construct deeper structures. Table \ref{tab:nas-stack-cmp} compares the results with 1 and 2 stacked cells. It is observer that deeper structures with 2 stacked cells gain no improvements, perhaps due to the gap between search and evaluation \cite{chen2019progressive} and overfitting. Perhaps, search with multiple stacked cells could boost the performance but it leads to heavy computation consumption. For simplicity, we use only 1 cell in our experiments.
\begin{table}[t]
    \centering
    \begin{tabular}{c|c|c}
        \hline
        Settings & V1 Val Acc & V2 Val Acc \\
        \hline\hline
        1 cells & 51.4 & 63.5 \\
        2 cells & 50.7 & 63.0 \\
        \hline
    \end{tabular}
    \caption{Interaction recognition accuracy (\%) comparison of stacking multiple computation cells.}
    \label{tab:nas-stack-cmp}
\end{table}

\section{Analysis of Graph Operations}
\subsection{More Successful and Failed Cases}
\begin{figure}[t]
    \centering
    \subfigure[Stuffing something into something]{
    \label{fig:supp-case-study-feat-agg}
    \includegraphics[width=0.24\linewidth]{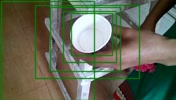}
    \includegraphics[width=0.24\linewidth]{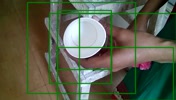}
    \includegraphics[width=0.24\linewidth]{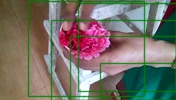}
    \includegraphics[width=0.24\linewidth]{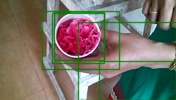}
    }
    \subfigure[Putting something that can't roll onto a slanted surface, so it slides down]{
    \label{fig:supp-case-study-diff-agg}
    \includegraphics[width=0.24\linewidth]{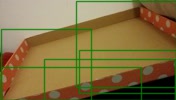}
    \includegraphics[width=0.24\linewidth]{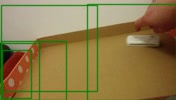}
    \includegraphics[width=0.24\linewidth]{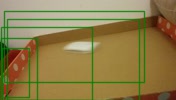}
    \includegraphics[width=0.24\linewidth]{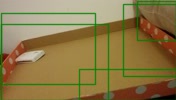}
    }
    \subfigure[Pushing something so that it slightly moves]{
    \label{fig:supp-case-study-tem-conv}
    \includegraphics[width=0.24\linewidth]{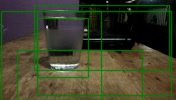}
    \includegraphics[width=0.24\linewidth]{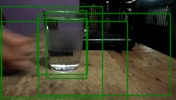}
    \includegraphics[width=0.24\linewidth]{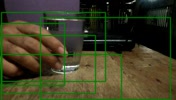}
    \includegraphics[width=0.24\linewidth]{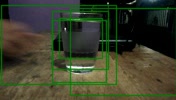}
    }
    \subfigure[Pushing something with something]{
    \label{fig:supp-case-study-back-rel}
    \includegraphics[width=0.24\linewidth]{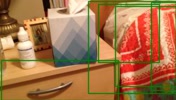}
    \includegraphics[width=0.24\linewidth]{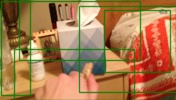}
    \includegraphics[width=0.24\linewidth]{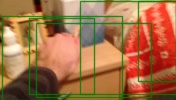}
    \includegraphics[width=0.24\linewidth]{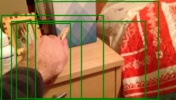}
    }
    \subfigure[Putting something on the edge of something so it is not supported and falls down]{
    \label{fig:supp-case-study-all-failed}
    \includegraphics[width=0.24\linewidth]{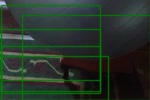}
    \includegraphics[width=0.24\linewidth]{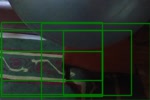}
    \includegraphics[width=0.24\linewidth]{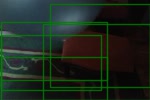}
    \includegraphics[width=0.24\linewidth]{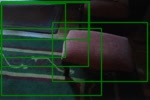}
    }
    \caption{Successful and failed cases of different graph operations.}
    \label{fig:supp-case-study}
\end{figure}

We show more successful and failed cases to indicate the effects of different operations in Figure \ref{fig:supp-case-study}.

The \textit{feature aggregation} successfully recognizes the simple interaction in (a), but it fails in case (b), (c), and (d) because some detailed relations need to be modeled with other operaiotns.

In case (b), the \textit{feature aggregation} easily recognizes the interaction as ``Letting something roll down a slanted surface''. The key to distinguish the two interactions is the differences between rolling and sliding. The \textit{difference propagation} focuses on the differences between detected objects, which enables it to capture the changes of the sliding down object so that it successfully classifies the interaction as sliding.

In case (c), the \textit{feature aggregation} mistakenly recognizes the interaction as ``Moving something and something closer to each other'' since the two objects are indeed closing. However, the key to identify is whether the two objects are both moved. The \textit{temporal convolution} aims to capture the evolution of the interaction and it could observe that one of the objects is always static, which makes the interaction distinguishable.

As for case (d), something is pushed but did not fall down but the \textit{feature aggregation} misclassifies it as ``Pushing something off of something''. The background would change dramatically if the box falls off, so the \textit{background incorporation} modeling the relations between the nodes and the background helps to identify the action. On the contrary, those operations relying on detected objects easily fail because the table is not detected by RPN model.

However, there are still many cases that are commonly misclassified by different graph operations and the searched structures, such as the example showed in Figure \ref{fig:supp-case-study-all-failed}. The poor quality frames prevent any effective modeling based on RGB inputs. What's more, some confusing labels and incorrect detected object bounding boxes also hinder further improvements of interaction modeling, which need to be addressed in the future.

\begin{figure}[t]
    \centering
    \subfigure[Something-Something-V1]{
    \label{fig:acc-cmp-V1}
    \includegraphics[width=1\linewidth]{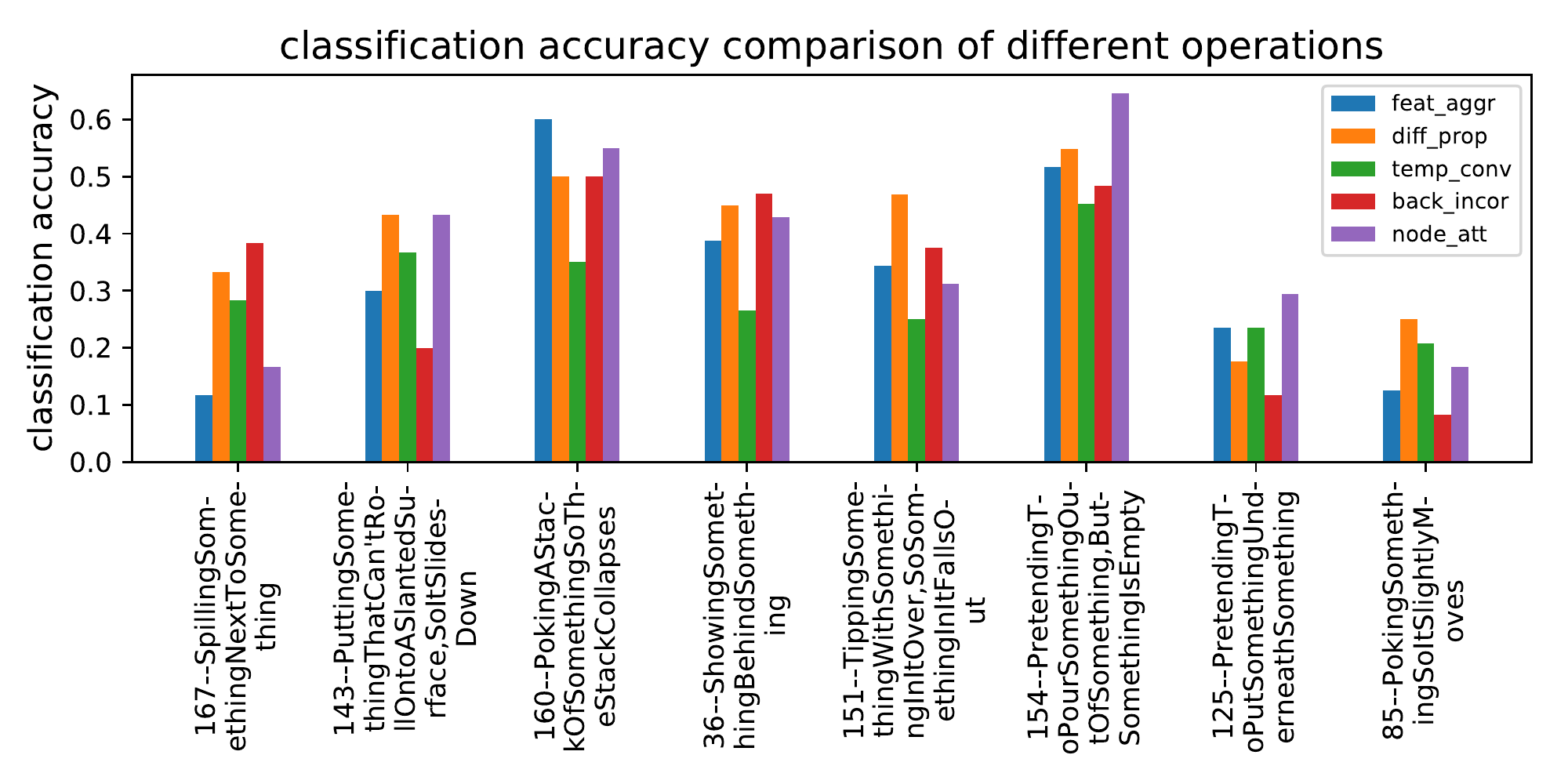}
    }
    \subfigure[Something-Something-V2]{
    \label{fig:acc-cmp-V2}
    \includegraphics[width=1\linewidth]{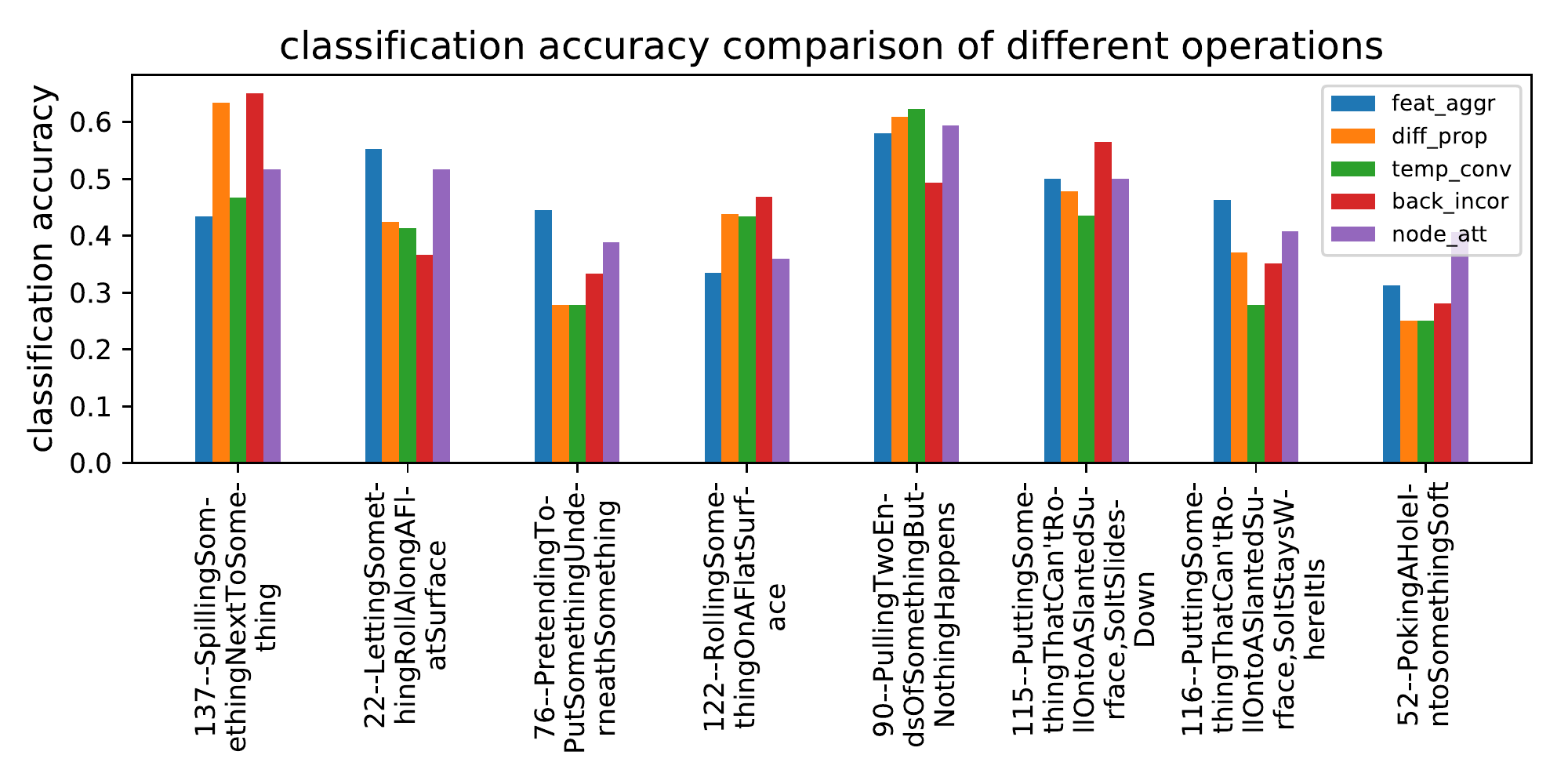}
    }
    \caption{Accuracy comparison of each graph operation on some interaction categories.}
    \label{fig:acc-cmp}
\end{figure}

\subsection{Accuracy of Each Graph Operation}
To show the effects of different graph operations on different interaction categories, we compare the recognition accuracy of each graph operation on some interaction categories where different operations obtain quite different performances. The results are shown in Figure \ref{fig:acc-cmp}. It is observed that different operations perform differently on the same interaction category, since they tend to model different relations in videos. We can also observe that the \textit{different propagation} and the \textit{temporal convolution} generally work well on the interactions with detailed changes, such as spilling something, moving something slightly and rolling something, and the \textit{background incorporation} would work well on some interactions with relations between different objects and the background, such as positional relations and relations to the surfaces.

\clearpage
{\small
\bibliographystyle{ieee_fullname}
\bibliography{egbib}
}

\end{document}